%% file: main.tex
\documentclass{article}

\usepackage{style_iclr2027/fancyhdr}
\usepackage{style_iclr2027/natbib}
\usepackage{style_iclr2027/iclr2027_conference,times}
\iclrfinalcopy

\usepackage[utf8]{inputenc} 
\usepackage[T1]{fontenc}    
\usepackage{url}            
\usepackage{booktabs}       
\usepackage{amsfonts}       
\usepackage{nicefrac}       
\usepackage{microtype}      
\usepackage{xcolor}         

\input{preamble}

\newcommand{\papertitle}{%
Vision Harnessing Agent for Open Ad-hoc Segmentation
}

\title{\textls[0]{Vision Harnessing Agent \\for Open Ad-hoc Segmentation}}

\author{%
Zilin Wang \qquad Stella X. Yu \\
University of Michigan \\
{\small code: \href{https://github.com/Wayne2Wang/VASA}{github.com/Wayne2Wang/VASA}}
}

\begin{document}

\maketitle

\lhead{Preprint.}

\etocdepthtag.toc{mtchapter}
\etocsettagdepth{mtchapter}{subsection}
\etocsettagdepth{mtappendix}{none}
\faketableofcontents

\input{sections/0_abstract}

\input{sections/1_intro}

\input{sections/2_related}

\input{sections/3_method}

\input{sections/5_experiments}

\input{sections/6_summary}

\subsection*{AI use statement}

We used Qwen3-VL 8B Instruct to construct our PARS benchmark, which the authors manually verified and corrected. (Sec.~\ref{sec:exp} and Appendix~\ref{appendix:pars}). We also used AI tools for manuscript writing, polishing, and presentation. The authors take responsibility for the final content and results.

\subsection*{Reproducibility statement}
We will publicly release our code and data to enable reproduction of all results reported in this paper. The full \sname inference algorithm and additional implementation settings are provided in \app~\ref{appendix:vasa} and \ref{appendix:implementation}.

{
\small
\bibliographystyle{style_iclr2027/iclr2027_conference}
\bibliography{main}
}

\input{sections/X1_appendix}

\end{document}

%% file: preamble.tex
\definecolor{citecolor}{HTML}{0071BC}
\definecolor{swgreen}{rgb}{0, 0.6823, 0.7215}
\definecolor{gaingreen}{rgb}{0.125,0.549,0.129}
\definecolor{lightgraytext}{gray}{0.6}
\definecolor{lightblue}{rgb}{0.0, 0.9, 1}

\usepackage{xspace}
\usepackage{graphicx}
\usepackage[pagebackref,breaklinks,colorlinks]{hyperref}
\hypersetup{
    colorlinks,
    linkcolor={red},
    citecolor={citecolor},
}
\usepackage{amsmath}
\usepackage{cleveref}
\usepackage{array}
\usepackage[table]{xcolor}
\usepackage{caption}
\usepackage{textcomp}
\usepackage{multirow}
\usepackage{pifont}
\usepackage[ruled,vlined]{algorithm2e}

\usepackage{minitoc}
\usepackage{etoc}
\etocsettocstyle{\section*{Contents}}{}

\newcommand{\Quotes}[1]{{\textit{``#1''}}}
\newcommand{\sota}[1]{{\textbf{#1}}}
\newcommand{\sotb}[1]{{\underline{#1}}}
\newcommand{\ign}[1]{\textcolor{lightgraytext}{#1}}

\newcommand{\cmark}{\color{gaingreen} \ding{51}}%
\newcommand{\xmark}{\color{red} \ding{55}}%
\newcommand{\fig}{Fig.\xspace}%
\newcommand{\app}{Appendix\xspace}%

\renewcommand{\cite}[1]{\citep{#1}}

\newcolumntype{a}{>{\columncolor{lightblue!10}[0pt][0pt]}c}
\newcolumntype{e}{@{\hspace{3pt}}c}
\newcolumntype{f}{@{\hspace{-3pt}}c}

\newcommand{\gaint}[1]{\textbf{{\color{gaingreen}#1}}}
\newcommand{\gain}[1]{\gaint{\makebox[0pt][r]{+}\hspace{0pt}#1}}
\newcommand{\gains}[1]{\gaint{+#1}}

\newcommand{\stask}{PARS\xspace}
\newcommand{\ltask}{\underbar{P}artImageNet \underbar{A}d-hoc \underbar{R}eferring \underbar{S}egmentation\xspace}
\newcommand{\sname}{VASA\xspace}
\newcommand{\lname}{\underbar{V}ision-guided \underbar{A}d-hoc \underbar{S}egmentation \underbar{A}gent\xspace}
\newcommand{\ours}{\sname (ours)\xspace}
\newcommand{\sam}{SAM\xspace}
\newcommand{\samtwo}{SAM2\xspace}
\newcommand{\samthree}{SAM3\xspace}
\newcommand{\samthreeagent}{SAM3 Agent\xspace}
\newcommand{\ssamthreeagent}{S.A.\xspace}
\newcommand{\refcocom}{RefCOCOm\xspace}

\newcommand{\lisa}{LISA\xspace}
\newcommand{\lisapp}{LISA++\xspace}
\newcommand{\resany}{RESAnything\xspace}
\newcommand{\partin}{PartImageNet\xspace}
\newcommand{\gres}{GRES\xspace}
\newcommand{\grefcoco}{gRefCOCO\xspace}
\newcommand{\segtoken}{\texttt{[SEG]}\xspace}
\newcommand{\miou}{gIoU\xspace}
\newcommand{\ciou}{cIoU\xspace}

\newcommand{\cpc}{xIoU\xspace}
\newcommand{\segagent}{SegAgent\xspace}
\newcommand{\cores}{CoReS\xspace}
\newcommand{\segzero}{Seg-Zero\xspace}
\newcommand{\visionreasoner}{VisionReasoner\xspace}
\newcommand{\ovseg}{OVSeg\xspace}
\newcommand{\semanticsam}{Semantic-SAM\xspace}
\newcommand{\vlpart}{VLPart\xspace}
\newcommand{\metrics}{\miou & \ciou & \cpc\!$\downarrow$}

\newcommand{\refcocommetrics}{Part & +Obj}
\newcommand{\oas}{open ad-hoc segmentation\xspace}

\newcommand{\reasonseg}{ReasonSeg\xspace}
\newcommand{\guidelineseg}{GuidelineSeg\xspace}
\newcommand{\evolsamthree}{Evol-SAM3\xspace}

%% file: sections/0_abstract.tex
\vspace{-6pt}
\begin{abstract}
\vspace{-6pt}

Segmentation has become easy when the concept is known, requiring retrieval of a learned visual grounding from text. It remains hard for open ad-hoc concepts, where the grounding may not exist as one learned mask and must often be constructed from image evidence through parts, relations, exclusions, and collections. We propose a \lname (\sname), the first vision harnessing agent for \oas. \sname is training-free and couples a VLM agent, a segmentation foundation model, and a visual harness that maintains a working mask to make visual progress persistent, inspectable, and editable. Rather than revising text prompts alone, it plans visual operations, invokes segmentation tools, inspects results, edits the mask, and recovers from errors. We construct \stask, a new benchmark that turns part-level labels into open ad-hoc concepts through long-form definition queries. We show that \sname is consistently effective across six VLMs with varying capabilities. Using Qwen3-VL 32B Thinking as the VLM, \sname outperforms various baselines on \stask, surpassing \samthreeagent by 13.5\%--25.3\%. On \refcocom, \sname improves over \samthreeagent by 4.8\%--8.8\% and over other agentic baselines by more. \sname also remains competitive with \samthreeagent on \reasonseg for common, named concepts. These results validate \sname's agentic visual construction for \oas. 

\end{abstract}

%% file: sections/1_intro.tex
\section{Introduction}
\label{sec:intro}

Segmentation has become easy when the concept is known. Modern foundation models can segment familiar visual wholes, such as {\it cats} and {\it cars}, because these concepts are visually coherent, commonly named, visually grounded, and richly represented in training data. \samthree~\cite{carion2026sam3} builds on the success of \sam~\cite{kirillov2023sam} and \samtwo~\cite{ravi2024sam2} by using large-scale text-region supervision, achieving strong performance across diverse segmentation benchmarks.  

However, segmentation remains hard when the concept is open and ad-hoc (\fig~\ref{fig:teaser}). A user may ask not for a named concept, but for an arbitrary one involving parts, relations, exclusions, or collections, constructed on the fly for the purpose at hand. We call this setting \textit{\oas}: Segmenting open-ended, task-contingent visual concepts beyond retrieval of established concepts. This setting extends {\it open ad-hoc categorization}~\cite{wang2025open}, which contextualizes recognition on demand, but imposes a stricter demand: The ad-hoc concept must be grounded at the pixel level.

\input{floats/fig_teaser}

Large-scale language and vision-language models (LLM/VLMs) have shown strong capabilities in text-conditioned visual reasoning, making it possible to pair segmentation foundation models with language-based reasoning beyond direct concept matching. One direction finetunes VLMs to produce intermediate segmentation representations, such as the \segtoken token in \lisa~\cite{lai2023lisa} and geometric prompts in \segzero~\cite{liu2025segzero}. Another direction uses VLM agents to decompose hard tasks into executable steps, treating an external segmenter as a tool. The agent interprets a complex query, proposes geometric prompts as in \segagent~\cite{zhu2025segagent} or text prompts as in \samthreeagent~\cite{carion2026sam3}, examines the resulting masks, and tries again. Their reasoning advances through revised prompts, but \emph{their visual solution does not build up accordingly} (\fig~\ref{fig:teaser},~\ref{fig:walkthrough}).

This disconnect between textual reasoning and visual progress hinders their performance on \oas, but does not necessarily imply that the models lack the required perceptual or reasoning capabilities. They may instead lack an effective harness to guide their use. This motivates our research question: \emph{How can we harness VLM agents for effective open ad-hoc segmentation?}

In this work, we introduce a {\it \lname} ({\bf \sname}), the first vision harnessing agent for \oas. \sname is a training-free framework that couples a VLM agent, a segmentation foundation model, and a visual harness. Unlike the previous state-of-the-art, \samthreeagent~\cite{carion2026sam3}, which retains textual reasoning history but not the evolving visual construction, \sname maintains a working mask that \emph{makes visual progress persistent, inspectable, and editable}. The agent can add newly found regions, remove incorrect regions, replace coarse masks with refined ones, and verify progress against the user query. Unlike \samthreeagent that are built with prompt and context engineering, \sname is built with a complete visual harness over state management, tool invocation, action constraints, planning, scrutiny, and error recovery. 

\fig~\ref{fig:walkthrough} compares the workflows of \sname and \samthreeagent. The side-by-side comparison reveals the key difference: \samthreeagent searches over prompts, advancing text reasoning without accumulating visual progress, whereas \sname incrementally constructs a visual solution. Text reasoning and visual construction proceed together, enabling \sname to successfully segment this open ad-hoc concept.

For evaluation, we construct \textbf{\stask}: {\it \ltask}, an \oas benchmark that turns part-level labels in \partin~\cite{he2022partimagenet} into open ad-hoc concepts through long-form definition queries. Starting from annotated examples, we prompt a VLM to write segmentation instructions that specify what to include, what to exclude, and how the concept is structurally or contextually identified; the generated queries are then manually verified. Instead of short, often ambiguous category names, \stask provides detailed, unambiguous queries. 

\input{floats/fig_walkthrough}

We first assess \sname's generalization across six VLMs spanning open-source and proprietary model families with varying capabilities. Our results show that \sname works effectively on older VLMs and scales well to the latest VLMs. On \stask, \sname consistently outperforms open-vocabulary, reasoning-based, and agentic segmentation baselines, surpassing \samthreeagent by 13.5\%-25.3\% using Qwen3-VL 32B Thinking~\cite{bai2025qwen3vl}. On the multi-granularity referring segmentation benchmark \refcocom~\cite{wang2024mres}, \sname outperforms \samthreeagent by 4.8\%-8.8\% and other agentic baselines by more. We further present ablation studies, additional analysis, qualitative results, and show that \sname is still on par with \samthreeagent on \reasonseg~\cite{lai2023lisa}, a standard reasoning segmentation benchmark on common, named concepts. 

\textbf{Contributions:} {\bf 1)} We introduce \oas, a challenging segmentation setting for open-ended, ad-hoc visual concepts. {\bf 2)} We develop \sname, the first vision harnessing agent for \oas, with persistent visual construction and complete visual harness. {\bf 3)} We curate \stask, a benchmark that uses detailed long-form queries to evaluate \oas. {\bf 4)} We show that \sname generalizes across VLMs, significantly outperforms the state of the art on \stask and \refcocom, and outperforms \samthreeagent on \reasonseg.

%% file: floats/fig_teaser.tex
\begin{figure*}[t]
\small\centering
\newcommand{\squareimage}[1]{\includegraphics[width=0.122\linewidth, height=0.103\linewidth]{#1}}
\newcommand{\colspace}{\hspace{8pt}}
\newcommand{\colspaceclose}{\hspace{1pt}}
\newcommand{\colspacenone}{\hspace{0pt}}
\newcommand{\tablerow}[2]{
\squareimage{figures/fig_teaser/#1.jpg} &
\squareimage{figures/fig_teaser/#1_prompt_#2} &
\squareimage{figures/fig_teaser/#1_lisa_#2} &
\squareimage{figures/fig_teaser/#1_segzero_#2} &
\squareimage{figures/fig_teaser/#1_segagent_#2} &
\squareimage{figures/fig_teaser/#1_sam3agent_#2} &
\squareimage{figures/fig_teaser/#1_ours_#2}
}
\renewcommand{\arraystretch}{0.3}
\begin{tabular}
{@{}c@{\colspacenone}c@{\colspace}
c@{\colspaceclose}c@{\colspaceclose}c@{\colspaceclose}
c@{\colspace}
c@{}}
Image & Prompt & \lisa & \segzero & \segagent & \samthreeagent & \ours \\ [2pt]
\tablerow{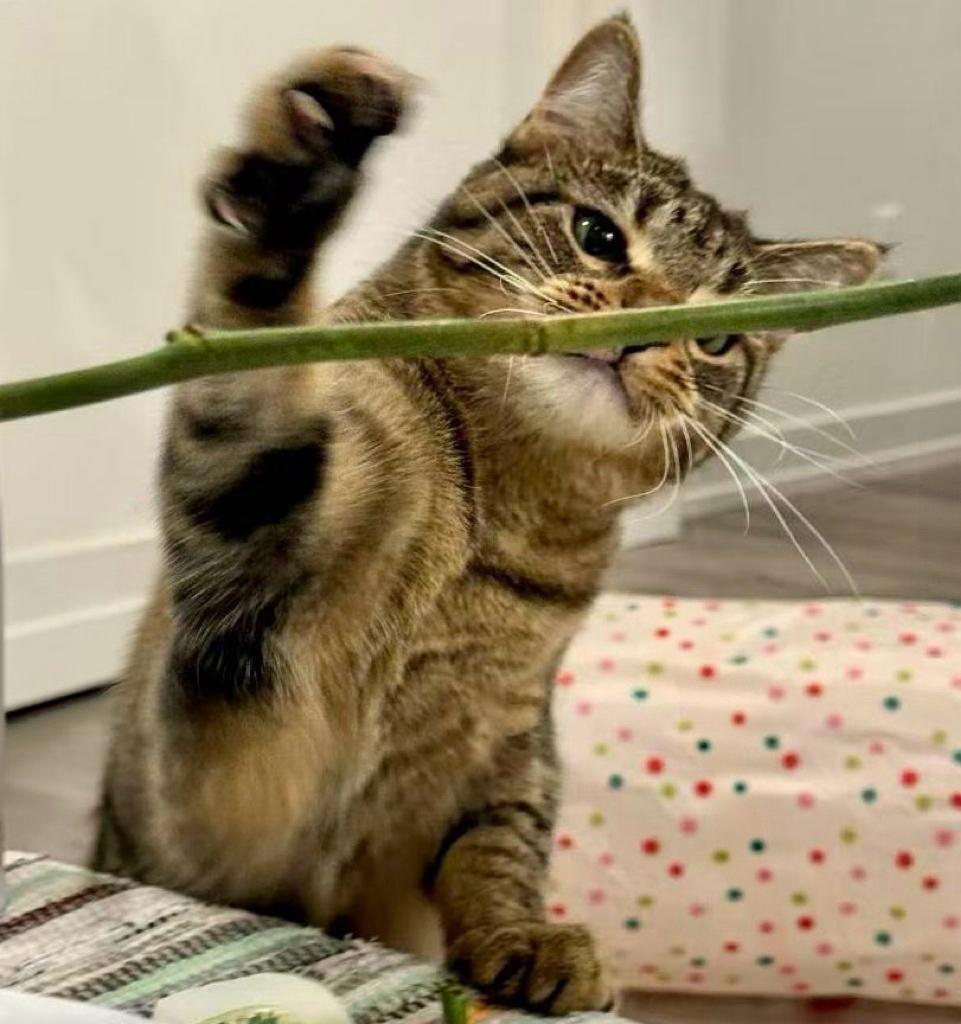}{1}\\
\tablerow{pipi}{2}\\
\tablerow{pipi}{3}\\
\end{tabular}
\caption{\textbf{Open ad-hoc segmentation requires constructing visual concepts on the fly, not merely retrieving common ones.}
We compare \lisa~\cite{lai2023lisa}, \segzero~\cite{liu2025segzero}, \segagent~\cite{zhu2025segagent}, \samthreeagent~\cite{carion2026sam3}, and \sname (our proposed vision harnessing agent),  on three prompts for the same image. \textbf{Row 1} asks for a known visual whole, {\it cat}; all methods succeed. \textbf{Row 2} asks for an ad-hoc collection, {\it the cat's right paw and the stick she is reaching for}; baselines retrieve only one part of the requested composition, while \sname captures both. \textbf{Row 3} asks for an exclusion-defined concept, {\it the cat's head without the ears and eyes}; baselines confuse it with {\it cat head} or {\it cat}, while \sname constructs the requested mask.
}
\label{fig:teaser}
\end{figure*}

%% file: floats/fig_walkthrough.tex
\begin{figure}[t]
    \centering
    \includegraphics[width=0.9\textwidth]{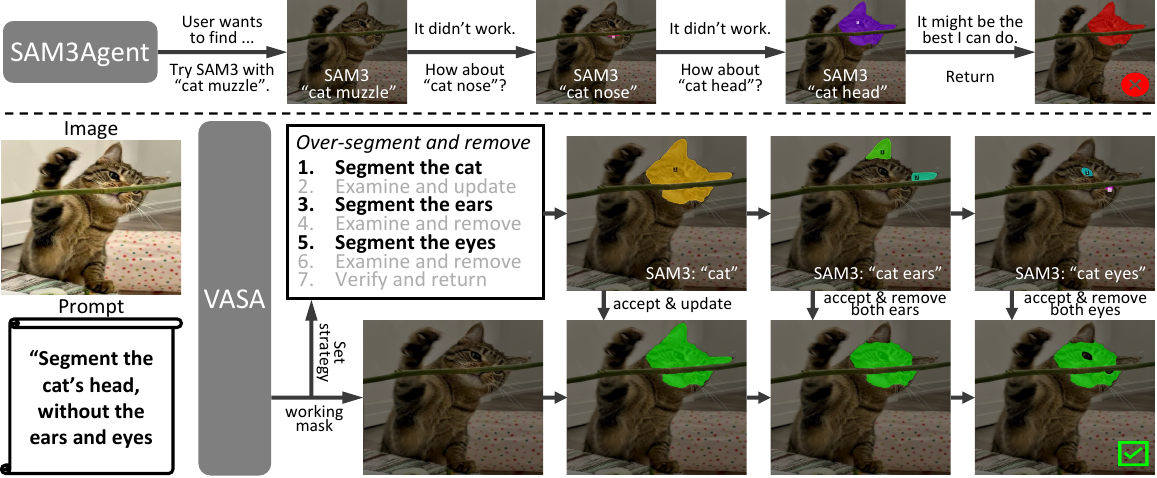}
    \caption{
\textbf{\sname reasons, constructs, and validates an open ad-hoc solution, while \samthreeagent searches over text prompts.} We consider the query {\it the cat's head, without the ears and eyes}. \textbf{Top:} \samthreeagent revises text queries to \samthree, trying prompts like {\it cat muzzle}. Each trial is a fresh retrieval attempt: Intermediate masks are not preserved for the final answer. \textbf{Bottom:} \sname instead plans a visual strategy, maintains a persistent working mask, and updates it via executable operations.
}
\vspace{-2pt}
\label{fig:walkthrough}
\end{figure}

%% file: sections/2_related.tex
\section{Related Work}
\label{sec:related}

\vspace{-1pt}

Additional related work on \textbf{referring segmentation} and \textbf{reasoning segmentation} is in \app~\ref{appendix:related}.

\textbf{Referring segmentation} focuses on segmenting image regions specified by explicit text descriptions. Much of its success comes from scaling region-text alignment supervision, like the semi-automatic data engine in \samthree~\cite{carion2026sam3}. Such scaling works well for visually coherent, commonly named objects, but cannot exhaust the space of open ad-hoc concepts users may define on demand.

\textbf{Reasoning segmentation} focuses on segmenting typically well-defined targets from \emph{implicit} text queries through commonsense and spatial reasoning. Most existing works use VLMs for reasoning and segmenters for mask decoding. Three types of reasoning paradigms prevail. \textbf{1)} Implicit reasoning, where the VLM predicts the target segment representation, keeping the reasoning process inaccessible, such as the \segtoken token in \lisa~\cite{lai2023lisa}. \textbf{2)} Text-based visual reasoning, where the VLM is tuned through RL~\cite{shao2024grpo} to reason in text, such as the target's coordinate prediction in \segzero~\cite{liu2025segzero}. \textbf{3)} Grounded visual chain-of-thought, where each reasoning step is visually grounded, and subsequent steps build on both prior reasoning and grounding results, such as SegLLM~\cite{wang2024segllm}. Open ad-hoc segmentation is an orthogonal direction, reasoning about what constitutes a visual concept rather than how to find the existing concept hinted by the user.

\textbf{Agentic segmentation} pairs VLMs with segmenters through \emph{iterative tool use}. Beyond reasoning segmentation, they also refine how the target is delineated or composed. Prior works like \cores~\cite{bao2024cores} decomposes segmentation into coarse-to-fine stages, while \segagent~\cite{zhu2025segagent}, RSAgent~\cite{he2025rsagent}, and SAM Veteran~\cite{du2026samveteran} use finetuned VLMs to iteratively refine geometric prompts. \evolsamthree~\cite{ye2025evolsam3} and \samthreeagent~\cite{carion2026sam3} instead operate without finetuning, using evolutionary search and sequential revision of text prompts, respectively. \guidelineseg~\cite{vats2025guidelineseg} considers a related setting in which the targets are defined by detailed guidelines, with a pipeline designed around only driving scenes that coordinates multiple pretrained models through Worker-Supervisor refinement and an RL-based stopping policy. Our \stask broadens it with a more diverse benchmark of object and part compositions. These methods approach \oas but remain limited by disconnected textual reasoning and visual progress. In contrast, \sname is the first vision harnessing agent to make visual progress persistent, inspectable, and editable, integrating other harness components for \oas.

\textbf{Harness engineering}~\cite{hashimoto2026aiadoption,openai2026harness} designs the full environment around an agent for reliable execution. It extends prompt and context engineering to govern actions, constraints, verification, and error recovery. Beyond software engineering, it has been used in visual games in VISTA~\cite{vista2026}, robot policy training in HARBOR~\cite{li2026harbor} and ENPIRE~\cite{xiao2026enpire}, and robot task execution in Thea~\cite{wang2026thea} and HarnessVLA~\cite{zhang2026harnessvla}. \sname is the first vision harnessing agent for \oas.

%% file: sections/3_method.tex
\section{Vision Harnessing Agent}
\label{sec:method}

\input{floats/fig_model}

Open ad-hoc segmentation calls for visual construction in which intermediate masks support further reasoning and refinement. We develop \sname (\lname) through a visual harness that preserves this visual progress and guides subsequent actions toward satisfying the user query, allowing reasoning and construction to advance together (\fig~\ref{fig:model}).

{\bf 1. Overview.}
\sname couples a VLM agent with a segmenter through a visual harness. The segmenter provides pixel-level grounding, returning candidate masks for the VLM's prompts. The VLM interprets the query, sets a strategy, chooses prompts, inspects candidate masks, and decides how to compose them into the target. The harness executes and constrains these decisions, maintains visual state, and guides planning, inspection, and recovery. Each result informs the VLM's next decision.

{\bf 2. State Management.}
Exploring new regions should not erase existing visual progress. We therefore maintain an initially empty working mask separately from the latest segmentation candidates. Segmentation calls refresh the candidates without committing them to the solution; only explicit edits change the working mask, which is rendered on the image after each edit for inspection. We retain the image, query, system instructions, strategy, and compact action history to help the agent track what has been achieved and what remains unresolved in the evolving construction.

{\bf 3. Tool Calls and Mask Editing.}
Visual construction requires translating the VLM's decisions into executable operations. We expose these operations through structured tool calls, spanning input analysis, strategy recording, segmentation, state updates, inspection, verification, and final output. Following \samthreeagent, \texttt{segment\_phrase} prompts \samthree with a short noun phrase and returns numbered candidate masks, while \texttt{examine\_each\_mask} supports closer inspection. Our \texttt{update\_working\_mask} executes VLM-selected \textsc{Add}, \textsc{Remove}, or \textsc{Replace} edits as deterministic pixel-level operations, adding or subtracting regions, or overwriting the working mask. Candidates thus need not match the entire query. Additional tools are introduced below.

{\bf 4. Long-Horizon Planning.}
Visual construction is a long-horizon planning problem in which intermediate masks may only partially satisfy the query. Setting a strategy beforehand gives these intermediate results a clear role and coordinates subsequent steps toward the target. We instruct the agent to begin with \texttt{set\_strategy}, analyzing the image and query to plan direct segmentation, undersegment-and-add, or oversegment-and-remove. The plan specifies initial prompts and subsequent edits, including preserving a base mask before seeking additions or exclusions. We retain the plan throughout inference to guide decisions alongside action history, tool reliability, progress, and the remaining budget, while allowing revisions when intermediate results reveal limitations.

{\bf 5. Constraints and Visual Scrutiny.}
Open ad-hoc segmentation requires deliberate observation and reasoning, supported by constraints that guide the VLM without prescribing every decision. We encourage close inspection before action by instructing the agent to examine fine-grained boundaries, attributes, and relations against the query, alongside the input image and working mask. For small or overlapping masks, \texttt{examine\_each\_mask} provides individual views for closer scrutiny. To carry these observations into actions, we design workflow instructions to guide reasoning, preserve accumulated state, restrict selection to current candidates, and discourage unnecessary replacement. We constrain tool use through structured output requirements and execution checks, leaving the VLM free to choose and revise strategies, prompts, and edits as visual evidence accumulates.

{\bf 6. Error Recovery and Stopping.}
In a long construction sequence, a failed step can mislead subsequent actions. We therefore design the harness to support recovery while preserving accumulated progress, rather than restarting the construction. When inspection reveals incorrect regions or stalled progress, \sname can revise its strategy, change prompts or concept granularity, and correct the working mask. Our execution checks return corrective feedback for repeated prompts, formatting errors, and invalid tool calls, enabling retries within the existing interaction. We instruct the agent to stop when the query is satisfied or further refinement appears unlikely to help, and impose a generation budget to limit execution. The harness returns the current working mask when available.

The \textbf{\sname inference algorithm} and \textbf{implementation details} are provided in \app~\ref{appendix:vasa} and \ref{appendix:implementation}.

%% file: floats/fig_model.tex
\begin{figure}[t]
    \centering
    \includegraphics[width=0.95\textwidth]{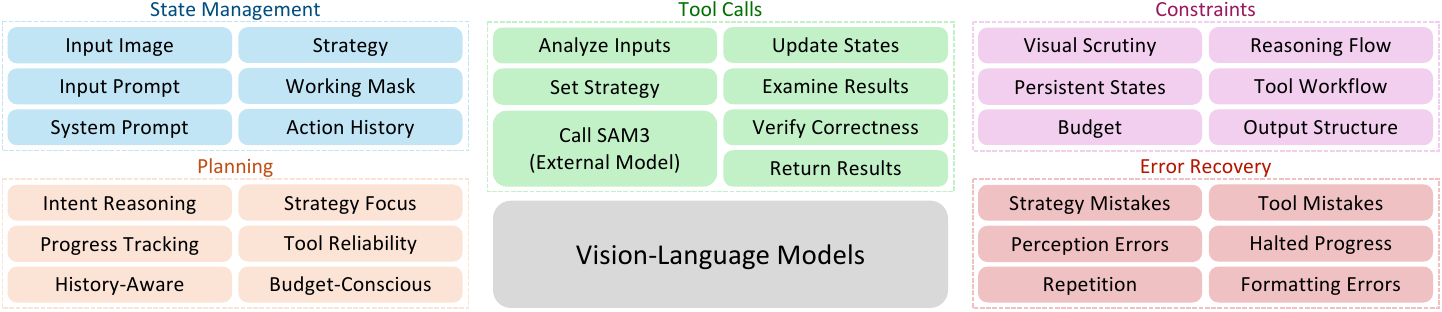}
    \caption{
    \textbf{\sname provides a complete visual harness around VLMs for \oas.} The five harness components form the full environment that supports steady visual progress, enabling complex open ad-hoc concepts to be progressively constructed from simpler visual primitives. The harness structures execution while leaving the VLM free to choose strategies, prompts, and edits. 
    }
    \label{fig:model}
\end{figure}

%% file: sections/5_experiments.tex
\section{Experiments on Open Ad-hoc Segmentation}
\label{sec:exp}

We evaluate \sname's visual harness on three settings: \textbf{1)} Our new \stask benchmark for \oas across six VLMs spanning different model families of various capabilities; \textbf{2)} The standard benchmark \refcocom for a related task of fine-grained multi-granularity referring segmentation; \textbf{3)} Ablation studies, further analysis, qualitative results, and results on \reasonseg.
 
\textbf{1. Datasets.}
\ltask (\textbf{\stask}) is a new benchmark we construct from \partin~\cite{he2022partimagenet} for open ad-hoc segmentation. \partin provides part annotations for different objects, such as the \Quotes{head/arm/foot/tail/body} of a \Quotes{gorilla} and the \Quotes{body/tire/side mirror} of a \Quotes{car}. We focus on the \Quotes{body} class because it is shared across objects but defined differently, making it ad-hoc and challenging. We replace the original short class labels with detailed long-form descriptions that precisely specify how the target concept should be constructed (Fig.~\ref{fig:instruction}). We then divide them into 937 \emph{open ad-hoc concepts} and 1,084 \emph{common concepts} across 2,021 images based on VLM's familiarity. We provide more details on \stask construction in Appendix~\ref{appendix:pars}.

\input{floats/fig_instruction}

We additionally evaluate on \refcocom~\cite{wang2024mres} and \reasonseg~\cite{lai2023lisa}. \refcocom is a multi-granularity referring segmentation benchmark containing both part- and object-level concepts. \reasonseg is a reasoning segmentation benchmark with implicit and long queries for typically otherwise common concepts. For both datasets, we use all validation and test splits and keep the original text prompts without further processing, unlike in \stask. This evaluation assesses whether our harness generalizes beyond open ad-hoc segmentation to common fine-grained part- and object-level localization and maintains its reasoning ability on common concepts.

\input{floats/tab_pin}

\textbf{2. Evaluation metrics. }
Following common practice, we adopt generalized Intersection-over-Union (gIoU) and cumulative Intersection-over-Union (cIoU), and introduce \textbf{\cpc} for evaluating \emph{cross-concept confusion} in datasets like \stask, where all relevant parts of objects are annotated.

Formally, given predicted mask $P_i$, ground-truth target mask $G_i$, and $O_i$, the union of all other annotated concept masks in the same image excluding the target concept, the metrics are defined as:
\begin{equation}
\text{gIoU} = \frac{1}{N}\sum_i \frac{|P_i \cap G_i|}{|P_i \cup G_i|},
\quad
\text{cIoU} = \frac{\sum_i |P_i \cap G_i|}{\sum_i |P_i \cup G_i|},
\quad
\text{\cpc} = \frac{1}{N}\sum_i \frac{|P_i \cap O_i|}{|P_i|}.
\end{equation}
Intuitively, \cpc measures how often predictions incorrectly include regions belonging to other close concepts in the same image. The \Quotes{x} in \cpc denotes both \Quotes{cross}-concept confusion and the \Quotes{error} rate from overlapping incorrect concepts. \cpc captures fine-grained mistakes that cause only small IoU drops but are critical for perfect segmentation and conceptual correctness. Since \cpc can be trivially reduced by conservative masks, we use it as a secondary metric alongside gIoU and cIoU.

{\bf 3. Experiment Settings.}
We consider three groups of baselines with increasing VLM involvement: 1) \emph{Open-vocabulary segmenters} for text-prompted segmentation without VLMs. 2) \emph{VLMs as Visual Reasoners}, which integrate segmenters into VLMs and are fine-tuned to reason before producing a segmentation mask. 3) \emph{Context agents for segmentation}, which involve more interaction between VLMs and segmenters and typically fine-tune VLMs to call segmenters iteratively as tools. We evaluate \sname with \samthree and six VLMs spanning open-source and proprietary families across various capabilities: Gemini 2.5 Flash Lite~\cite{comanici2025gemini25}, Qwen3-VL 32B Thinking~\cite{bai2025qwen3vl}, GPT-5.6 Luna~\cite{openai2026gpt56}, GPT-6 Luna~\cite{openai2026gpt6}, GLM 5.3 Flash~\cite{glm5team2026glm5vibecodingagentic}, and DeepSeek V4.1 Flash~\cite{deepseekai2026deepseekv41flash}, all using the same harness. We evaluate our closest competitor, \samthreeagent, with the same VLMs to make the comparison \emph{independent of VLM choice}. We use the open-source Qwen model for ablation studies and analysis. We also train a supervised segmenter, \vlpart, on the training set as a closed-world reference.

{\bf 4. Results on \stask.}
We evaluate \sname and the baselines on the \stask benchmark for \oas (Table~\ref{table:pin}). Our results show that \emph{the effectiveness of a segmentation agent depends strongly on how it is harnessed}. Across all six VLMs, \samthreeagent's performance saturates within a narrow range of 39.6\%-40.8\% ad-hoc gIoU. This suggests that, even with stronger VLMs, \samthreeagent's design fundamentally limits its ability to segment rarer and more complex concepts, leading it to confuse common concepts with the open ad-hoc concepts specified by the query. In contrast, \sname with the oldest VLM, Gemini 2.5 Flash Lite, already outperforms \samthreeagent with the latest DeepSeek V4.1 Flash by 7.2\%. This gap grows to 19.1\% when \sname uses the same DeepSeek V4.1 Flash backbone. The reduction in ad-hoc \cpc from 59.0\% to 30.4\% further shows that \sname more clearly distinguishes the requested concepts from related but incorrect regions and excludes those regions from its predictions. Our visual harness enables \sname to generalize more effectively to stronger VLMs, translating greater VLM capability into better downstream performance. These gains extend to the broader set of baselines, including a supervised segmenter trained directly on the benchmark's training set when using DeepSeek V4.1 Flash without any task-specific fine-tuning.

{\bf 5. Results on \refcocom.}
We further evaluate \sname on a related task of \emph{fine-grained multi-granularity referring segmentation} on \refcocom (Table~\ref{table:refcocom}). Unlike our \stask for \oas, \refcocom uses shorter referring expressions than \stask, such as \emph{the left leg of the man facing us}, which we use without modifications. Compared with all baselines, \sname achieves the best performance across all splits and evaluation settings by large margins. Compared with \samthreeagent, it improves gIoU by up to 8.8\% in the part-only setting and 5.9\% in the combined part-and-object setting. The gains are consistently larger in the part-only setting, where fine-grained targets can be confused with neighboring regions. Such targets call for the same capabilities emphasized by our harness: maintaining visual state, inspecting intermediate masks, and correcting errors before returning the final prediction. The improvements across both conventional referring segmentation methods and agentic baselines demonstrate the broader effectiveness of \sname's visual harness.

\input{floats/tab_refcocom}

{\bf 6. Harness component ablation.}
Table~\ref{tab:app-harness-ablation} examines how the harness components jointly support \sname's effective visual construction. The \emph{working-mask-only} variant maintains a persistent mask and allows the agent to update and inspect it, but lacks support from other harness components. This gives the agent the mechanism for accumulating visual progress without the guidance to use it reliably, improving \cpc on \stask from 48.0\% to 37.5\% but dropping the gIoU from 46.4\% to 40.0\%. Adding strategy partially addresses this limitation, raising gIoU to 44.9\%. Constraints and scrutiny further improve gIoU to 48.3\%, while the complete harness reaches 56.9\% with the lowest \cpc of 22.9\%. These results show that a working mask alone does not establish an effective visual construction process; it benefits from coordinated guidance from \sname's complete harness.

\input{floats/tab_harness_ablation}

{\bf 7. Which VLM capabilities matter most for \sname?}
Visual construction requires both \emph{multimodal understanding} and \emph{broader knowledge and reasoning}. Fig.~\ref{fig:vlm_capability_cost}(a,b) relates \sname's overall \stask gIoU to MMMU-Pro and HLE scores across different VLMs. GPT-6 Luna scores higher on HLE but lower on MMMU-Pro than GPT-5.6 Luna. On \stask, it improves \emph{common}-concept segmentation but leaves \emph{ad-hoc} gIoU unchanged and increases cross-concept confusion (\cpc\!$\downarrow$). This suggests that broader reasoning gains may not compensate for weaker multimodal understanding when grounding ad-hoc concepts. Both benchmarks correlate strongly with gIoU with Pearson $r=0.88$, re-verifying that \sname translates stronger VLM capabilities into better \oas performance.

\input{floats/fig_vlm_capability_cost}

{\bf 8. Effect of detailed queries for \oas.}
In \stask, we replace the original class names from \partin with detailed queries because short names often leave the intended open ad-hoc concepts ambiguous. Here, we revert to the original names while keeping the images and target masks unchanged to examine the effect of query detail. When using Qwen3-VL 32B Thinking as the VLM, detailed queries improve \sname's gIoU from 51.0\% to 56.9\% and reduce \cpc from 41.7\% to 22.9\%, whereas \samthreeagent's gIoU drops from 46.4\% to 24.3\% (Table~\ref{table:abinstruct}). Fig.~\ref{fig:abinstructexample} illustrates this distinction. This contrast shows that additional query detail benefits segmentation only when the agent can effectively act on it, and also confirms the need for detailed queries in \stask.

\input{floats/tab_ablation_instruction}

{\bf 9. Reasoning steps, token usage, and inference cost.}
We examine whether \sname's gains come simply from a \emph{larger inference budget} or from \emph{more effective use of computation}. Both \sname and \samthreeagent are allowed up to 20 reasoning rounds. Yet \samthreeagent stops after just 3.8 rounds on average despite its lower segmentation results, struggling to turn the remaining budget into further improvements. \sname instead uses 7.9 rounds, while making only slightly more segmentation calls (3.4 versus 3.1; Table~\ref{tab:runtime-main}). These additional rounds support planning, mask inspection, and refinement of a persistent visual solution, rather than repeated attempts to obtain the complete target from a single segmentation call. Consistent with this interpretation, larger increases in reasoning steps over \samthreeagent are associated with larger gIoU gains (Fig.~\ref{fig:reasonvsiou}). Although average runtime rises from 5.2 to 9.3 seconds per image, the additional computation supports productive visual construction, enabling \sname to make better use of the same available budget.

Across VLM backbones, however, better segmentation is often obtained using fewer tokens ($r=-0.68$; Fig.~\ref{fig:vlm_capability_cost}(c)), suggesting that more capable models can reach better solutions with fewer tokens. API cost has a less clear relationship with performance, as it also reflects provider-specific pricing (Fig.~\ref{fig:vlm_capability_cost}(d)). Among the evaluated configurations, GLM 5.3 Flash and GPT-6 Luna offer the strongest cost-performance tradeoffs, while DeepSeek V4.1 Flash achieves the highest gIoU at a relatively higher cost than the other models but still remains low compared to other frontier models.

\input{floats/fig_reason_vs_iou}

\input{floats/fig_qualitative}

{\bf 10. Qualitative results.}
Fig.~\ref{fig:qualitative} shows how \sname follows concept boundaries beyond familiar object masks. On \stask, \samthreeagent includes excluded regions, whereas \sname more closely isolates the requested \emph{taxi body}, \emph{shark trunk}, and \emph{dog torso}. On \refcocom, it better separates the specified \emph{legs}, \emph{torso}, or \emph{arm from the surrounding person}. These examples illustrate more faithful grounding of both detailed definitions and short referring expressions. Additional examples are provided in \app~\ref{appendix:qualitative}.

{\bf 11. Results on \reasonseg.}
With Qwen3-VL 32B Thinking, \sname achieves 72.1\% validation and 70.6\% test gIoU, compared with 71.9\% and 69.4\% for \samthreeagent. While intended for challenging open ad-hoc concepts, \sname remains competitive on conventional reasoning segmentation.

%% file: floats/fig_instruction.tex
\begin{figure}[hp]
    \centering
    \includegraphics[width=\textwidth]{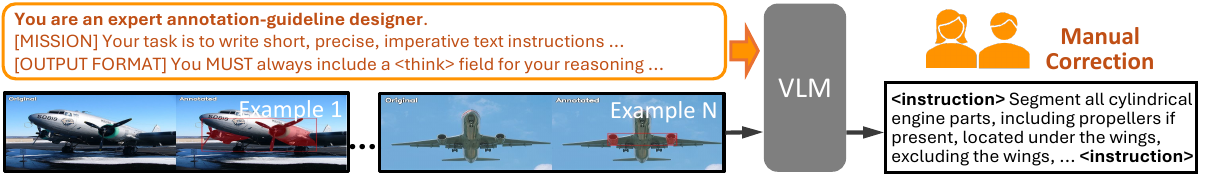}
\caption{\textbf{\stask provides detailed long-form descriptions for open ad-hoc concepts via a semi-automatic pipeline.} Given a small set of annotated examples for each concept, we prompt a VLM to disambiguate the target concept and what to include or exclude, followed by manual correction.}
    \label{fig:instruction}
\end{figure}

%% file: floats/tab_pin.tex
\begin{table*}[t!]

\caption{\textbf{\sname consistently outperforms prior methods across VLMs and remains competitive with a fully supervised segmenter on \stask.} All results are reproduced by the authors. We use \cpc as the secondary metric for cross-concept confusion, alongside \miou and \ciou. $^{*}$\semanticsam relies on a ground-truth-derived point prompt. $^{\dagger}$\guidelineseg uses Gemma 3 4B as a second VLM.}
\label{table:pin}
\newcommand{\close}{\hspace{3pt}}
\begin{minipage}{\linewidth}
\centering\small
\resizebox{1.0\linewidth}{!}{
\renewcommand{\arraystretch}{0.85}
\begin{tabular}[b]{ll@{\close}c c@{\close}c@{\close}a c@{\close}c@{\close}a c@{\close}c@{\close}a}
\toprule
\multirow{2}{*}{Method\vspace{-4pt}} & \multirow{2}{*}{VLM Version\vspace{-4pt}} & \multirow{2}{*}{\shortstack{Train. \\ Free}\vspace{-4pt}} & \multicolumn{3}{c}{Ad-hoc Concepts} & \multicolumn{3}{c}{Common Concepts} & \multicolumn{3}{c}{Total}\\
\cmidrule(lr){4-6} \cmidrule(lr){7-9} \cmidrule(lr){10-12}
& & & \metrics & \metrics & \metrics\\
\midrule
\multicolumn{4}{l}{\ign{\textit{Fully Supervised Segmenter}}} \\[3pt]
\ign{\vlpart~\cite{peize2023vlpart}}      & \ign{N/A} & \ign{N/A}  & \ign{55.2} & \ign{53.8} & \ign{35.1} & \ign{60.5} & \ign{73.3} & \ign{10.7} & \ign{57.5} & \ign{63.2} & \ign{23.9} \\
\midrule
\multicolumn{4}{l}{\textit{Open-Vocabulary Segmenter}} \\[3pt]
\ovseg~\cite{liang2023ovseg}                           & N/A       & N/A       & 34.4 & 27.6 & 42.2 & 46.0 & 43.0 & 20.7 & 39.3 & 34.2 & 33.5 \\
\semanticsam\!$^{*}$ ~\cite{li2023semanticsam}  & N/A       & N/A       & 39.2 & 30.4 & 45.2 & 51.7 & 53.7 & 25.7 & 44.5 & 39.4 & 38.4 \\
\samthree~\cite{carion2026sam3}                        & N/A       & N/A        & 34.1 & 36.1 & 59.8 & 41.8 & 57.8 & 30.6 & 37.3 & 45.0 & 48.9 \\
SAM3-I~\cite{li2026sam3i}                        & N/A       & N/A        & 45.8 & 47.1 & 26.2 & 38.3 & 46.0 & 30.4 & 42.6 & 46.5 & 28.4 \\
\midrule
\multicolumn{4}{l}{\textit{VLM as Visual Reasoner}} \\[3pt]
\lisa~\cite{lai2023lisa}                        &  Llama2 13B    & \xmark & 37.9 & 39.9 & 45.0 & 49.0 & 59.7 & 19.2 & 42.6 & 49.4 & 32.9 \\
\lisapp~\cite{yang2024lisapp}                   &  LLaVA1.5 7B   & \xmark & 35.2 & 33.7 & 52.6 & 51.8 & 50.4 & 25.4 & 42.3 & 41.5 & 41.0 \\
\visionreasoner~\cite{liu2026visionreasoner}    &  Qwen2.5-VL 7B & \xmark & 27.2 & 28.8 & 47.9 & 29.4 & 41.9 & 16.5 & 28.1 & 35.0 & 34.5 \\
\segzero~\cite{liu2025segzero}                  &  Qwen2.5-VL 7B & \xmark & 40.8 & 39.6 & 58.2 & 57.2 & 66.9 & 30.3 & 47.8 & 50.8 & 47.4 \\
\resany~\cite{wang2025resanything}              &  Qwen2.5-VL 7B & \cmark & 33.1 & 34.0 & 49.7 & 45.2 & 51.0 & 22.8 & 38.2 & 41.8 & 38.1 \\
\midrule
\multicolumn{4}{l}{\textit{Context Agent for Segmentation}} \\[3pt]
\cores~\cite{bao2024cores}             &  LLaVA 7B      & \xmark & 33.0 & 34.1 & 36.2 & 42.2 & 53.4 & 19.7 & 36.9 & 44.0 & 27.5 \\
\segagent\!-SC~\cite{zhu2025segagent}     &  Qwen-VL 7B    & \xmark & 34.6 & 34.8 & 46.6 & 48.2 & 61.8 & 20.9 & 40.4 & 47.9 & 33.8 \\
\segagent\!-\sam~\cite{zhu2025segagent}   &  Qwen-VL 7B    & \xmark & 33.1 & 31.7 & 48.2 & 45.9 & 56.0 & 19.8 & 38.5 & 43.1 & 35.0 \\
\guidelineseg\!$^{\dagger}$~\cite{vats2025guidelineseg}     &  Gemini 2.5 Flash & \xmark & 33.4 & 33.8 & 46.5 & 46.5 & 56.3 & 19.7 & 39.0 & 44.1 & 34.6 \\
\evolsamthree~\cite{ye2025evolsam3}     &  Qwen2.5-VL 7B  & \cmark & 41.4 & 44.4 & 43.9 & 50.7 & 57.1 & 22.3 & 45.3 & 50.3 & 35.0 \\
\arrayrulecolor{black!20}
\midrule[0.2pt]
\arrayrulecolor{black}
\multirow{6}{*}{\samthreeagent~\cite{carion2026sam3}}     &  Gemini 2.5 Flash Lite  & \cmark & 39.6 & 38.8 & 58.6 & 54.0 & 60.9 & 30.6 & 45.7 & 47.7 & 48.4 \\
     &  Qwen3-VL 32B Think & \cmark & 40.5 & 39.1 & 58.8 & 55.0 & 62.1 & 27.5 & 46.4 & 48.1 & 48.0 \\
     &  GPT-5.6 Luna  & \cmark & 40.8 & 39.2 & 58.8 & 57.2 & 68.3 & 27.4 & 47.7 & 50.7 & 47.3 \\
     &  GPT-6 Luna  & \cmark & 40.0 & 39.1 & 58.7 & 55.7 & 65.5 & 28.2 & 46.7 & 49.7 & 47.5 \\
     &  GLM 5.3 Flash  & \cmark & 40.5 & 39.5 & 58.4 & 57.3 & 67.3 & 26.7 & 47.6 & 50.7 & 46.7 \\
     &  DeepSeek V4.1 Flash  & \cmark & \sotb{40.7} & \sotb{39.2} & \sotb{59.0} & \sotb{57.3} & \sotb{68.2} & \sotb{26.5} & \sotb{47.7} & \sotb{50.6} & \sotb{47.3} \\
\midrule
\multicolumn{4}{l}{\textit{Vision Harnessing Agent for Segmentation}} \\[3pt]
\multirow{6}{*}{\textbf{\ours}}     &  Gemini 2.5 Flash Lite  & \cmark & 48.0 & 51.6 & 39.7 & 58.5 & 75.5 & 11.0 & 52.4 & 62.6 & 27.4 \\
     &  Qwen3-VL 32B Think & \cmark & 54.0 & 56.9 & 33.5 & 60.8 & 77.0 & 10.0 & 56.9 & 66.5 & 22.9 \\
     &  GPT-5.6 Luna  & \cmark & 54.6 & 56.5 & 37.2 & 61.8 & 73.9 & 11.6 & 57.7 & 64.5 & 26.5 \\
     &  GPT-6 Luna  & \cmark & 54.5 & 54.9 & 40.5 & 67.4 & 78.0 & 12.5 & 60.0 & 65.3 & 28.8 \\
     &  GLM 5.3 Flash  & \cmark & 55.0 & 57.8 & 36.3 & 65.5 & 78.6 & 11.2 & 59.5 & 67.5 & 25.4 \\
     &  DeepSeek V4.1 Flash  & \cmark & \sota{59.8} & \sota{62.7} & \sota{30.4} & \sota{68.6} & \sota{80.6} & \sota{10.3} & \sota{63.5} & \sota{71.4} & \sota{21.2} \\
\multicolumn{2}{l}{\gaint{\textit{largest gain v.s. prior SOTA} (\samthreeagent)}} &    & \gain{19.1} & \gain{23.5} & \gain{28.6} & \gain{11.3} & \gain{12.4} & \gain{16.2} & \gain{15.8} & \gain{20.8} & \gain{26.1} \\
\bottomrule
\end{tabular}
}
\end{minipage}
\end{table*}

%% file: floats/tab_refcocom.tex
\begin{table}[t]
\centering\small
\newcommand{\close}{\hspace{4pt}}
\newcommand{\sep}{\hspace{6pt}}
\renewcommand{\arraystretch}{0.8}
\begin{minipage}[t]{0.54\linewidth}
\vspace{0pt}
\centering
\begin{tabular}[b]{@{}lc@{\close}c c@{\close}c c@{\close}c@{}}
\toprule
\multirow{2}{*}{Method\vspace{-4pt}} & \multicolumn{2}{c}{val} & \multicolumn{2}{c}{testA} & \multicolumn{2}{c}{\phantom{0}testB}\\ [-1pt]
\cmidrule(lr){2-3} \cmidrule(lr){4-5} \cmidrule(l){6-7}
& \refcocommetrics & \refcocommetrics & \refcocommetrics\\ [-1pt]
\midrule
X-Decoder                       & 16.2 & 29.5 & 13.6 & 23.6 & 20.3 & 33.8 \\
SEEM                                & 16.1 & 29.4 & 13.6 & 23.4 & 20.4 & 33.9 \\
UniRES                             & 19.6 & 34.3 & 16.4 & 27.8 & 25.2 & 41.7 \\
\samthree                        & 12.3 & 17.3 & 9.0  & 13.0 & 20.7 & 26.1 \\
SAM3-I                        & 17.0 & 29.5 & 14.7 & 24.7 & 22.3 & 36.1 \\
\midrule
\lisa                               & 21.3 & 34.3 & 18.5 & 28.6 & 25.7 & 40.1 \\
GSVA                                & 11.4 & 23.1 & 9.2  & 19.2 & 16.8 & 28.2 \\
GLaMM                          & 21.4 & 35.3 & 18.6 & 29.5 & 26.9 & 41.1 \\
M$^2$SA                             & 22.4 & 35.5 & 19.9 & 30.1 & 27.1 & 41.4 \\
\resany                     & 27.6 & -    & 26.5 & -    & 25.8 & - \\
\midrule
\cores                              & 21.2 & 29.5 & 17.3 & 24.0 & 20.6 & 30.6 \\
\segagent\!-SC                  & 19.7 & 33.4 & 16.8 & 27.3 & 22.9 & 38.0  \\
\segagent\!-\sam  & 19.7 & 32.9 & 16.9 & 27.1 & 22.6 & 36.7  \\
\evolsamthree                   & 24.3 & 34.0 & 20.6 & 28.8 & 30.8 & 41.0 \\
\samthreeagent                   & \sotb{40.2} & \sotb{48.1} & \sotb{34.4} & \sotb{41.5} & \sotb{41.3} & \sotb{50.4}  \\
\midrule
\textbf{\ours}                                         & \sota{45.0} & \sota{51.1} & \sota{43.2} & \sota{47.4} & \sota{47.6} & \sota{54.1}  \\
\gaint{\textit{v.s. prior SOTA}}                       & \gains{4.8} & \gains{3.0} & \gains{8.8} & \gains{5.9} & \gains{6.3} & \gains{3.7} \\
\bottomrule
\end{tabular}
\end{minipage}%
\hfill
\begin{minipage}[t]{0.44\linewidth}
\caption{
\textbf{\sname achieves state-of-the-art performance on the standard multi-granularity referring segmentation benchmark \refcocom}, with the largest gains on the fine-grained parts. We report gIoU for part-only (Part) and combined part-and-object (+Obj), with Qwen3-VL 32B Thinking as the VLM. We compare \sname against \emph{open-vocabulary segmenters}, \emph{VLMs for visual reasoning}, and \emph{context agents}, with additional works: X-Decoder~\cite{zou2022xdecoder}, SEEM~\cite{zou2023seem}, GSVA~\cite{xia2024gsva}, GLaMM~\cite{hanoona2023GLaMM}, M$^2$SA~\cite{jang2025mmr}. We reproduce results for \samthree, SAM3-I, and all context agent methods in the third group; the remaining results are reported from \citet{wang2025resanything} and \citet{jang2025mmr}.
}
\label{table:refcocom}
\end{minipage}
\end{table}

%% file: floats/tab_harness_ablation.tex
\begin{table}[t]
\centering\small
\newcommand{\close}{\hspace{4pt}}
\begin{minipage}[t]{0.56\linewidth}
\vspace{0pt}\centering
\setlength{\tabcolsep}{3pt}\renewcommand{\arraystretch}{1.1}
\begin{tabular}{@{}l@{\close}c@{\close}c@{\close}c@{\close}c@{\close}c@{\close}c@{\close}a@{}}
\toprule
Method & W.M. & Strategy & C\&S & E.R. & \metrics \\
\midrule
\samthreeagent & -- & -- & -- & -- & 46.4 & 48.1 & 48.0\\
\midrule[0.1pt]
W.M. Only & \cmark & -- & -- & -- & 40.0 & 46.1 & 37.5\\
+ Strategy & \cmark & \cmark & -- & -- & 44.9 & 55.1 & 29.3\\
+ C\&S  & \cmark & \cmark & \cmark & -- & 48.3 & 59.9 & 27.3\\
\textbf{\ours} & \cmark & \cmark & \cmark & \cmark & \textbf{56.9} & \textbf{66.5} & \textbf{22.9}\\
\bottomrule
\end{tabular}%
\end{minipage}\hfill
\begin{minipage}[t]{0.42\linewidth}
\vspace{0pt}
\caption{
\textbf{Harness component ablation on \stask.} We start with working masks only and progressively add to the preceding setting. We use Qwen3-VL 32B Thinking and report \samthreeagent here as a reference. W.M.: working mask; C\&S: constraints \& scrutiny; E.R.: error recovery.
}
\label{tab:app-harness-ablation}
\end{minipage}
\end{table}

%% file: floats/fig_vlm_capability_cost.tex
\begin{figure}[t]
\centering
\begin{minipage}[t]{0.24\linewidth}
\centering
\includegraphics[width=\linewidth]{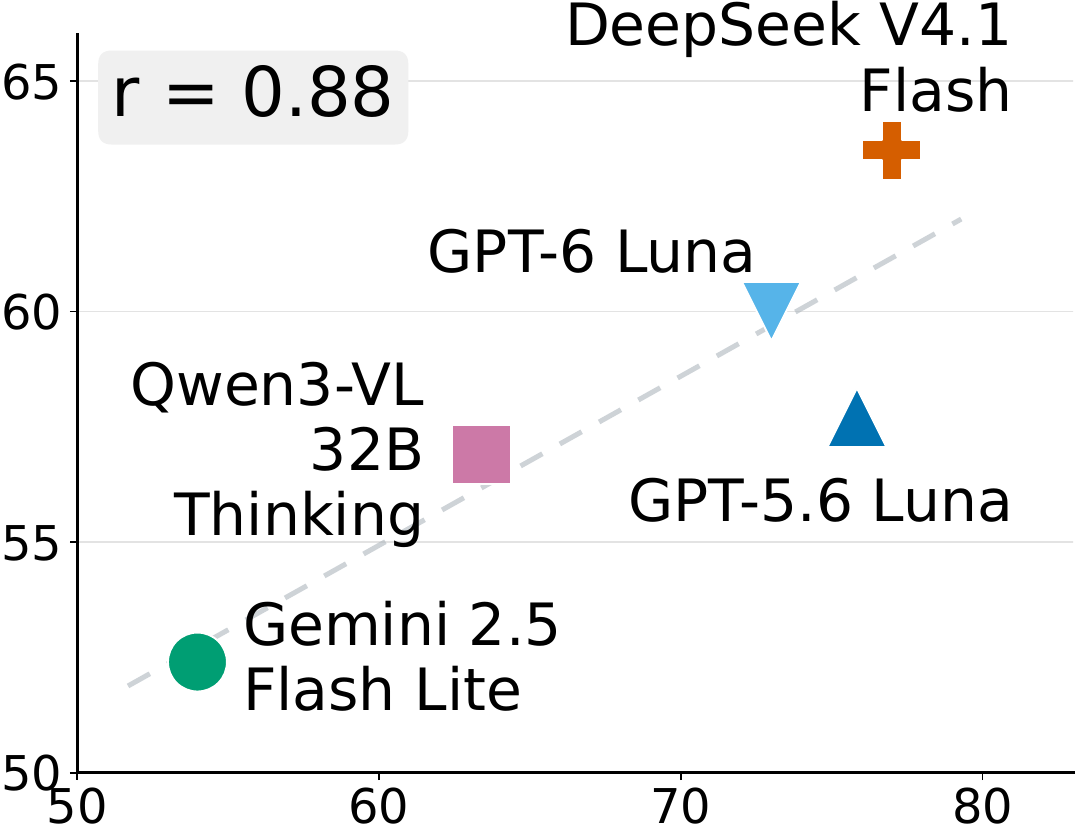}
\par\small a) MMMU-Pro score
\end{minipage}%
\hfill
\begin{minipage}[t]{0.24\linewidth}
\centering
\includegraphics[width=\linewidth]{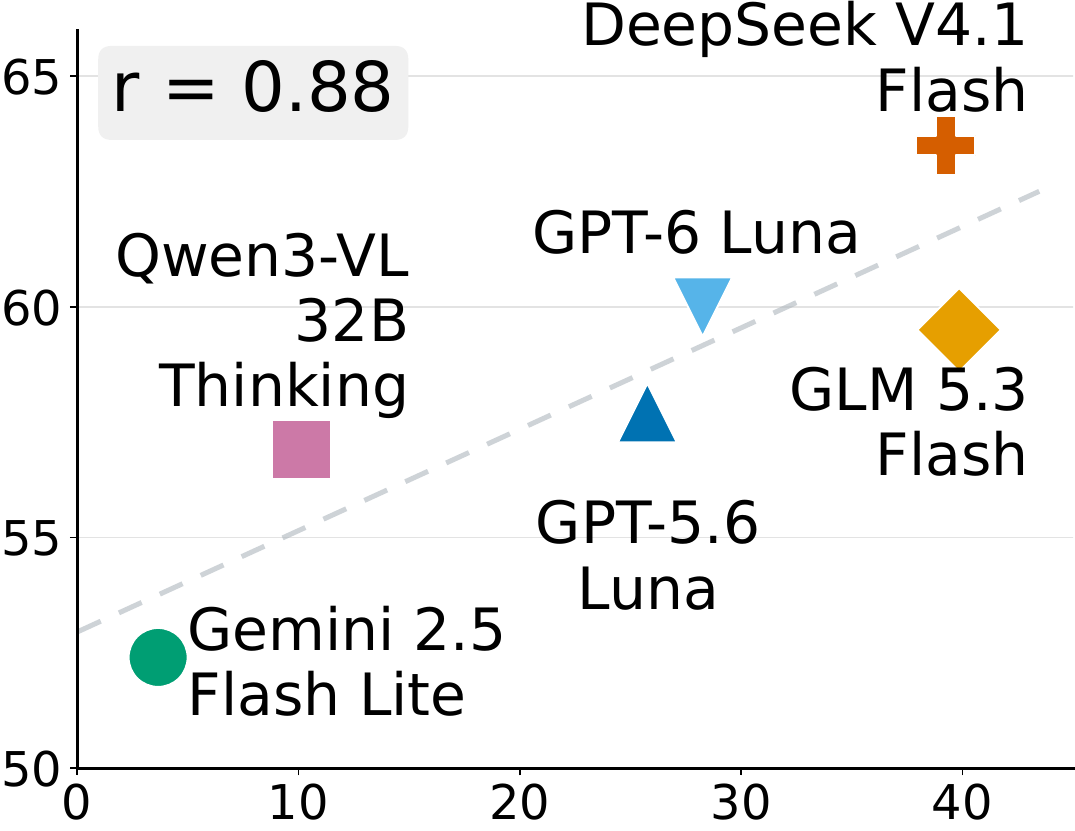}
\par\small b) HLE score
\end{minipage}%
\hfill
\begin{minipage}[t]{0.24\linewidth}
\centering
\includegraphics[width=\linewidth]{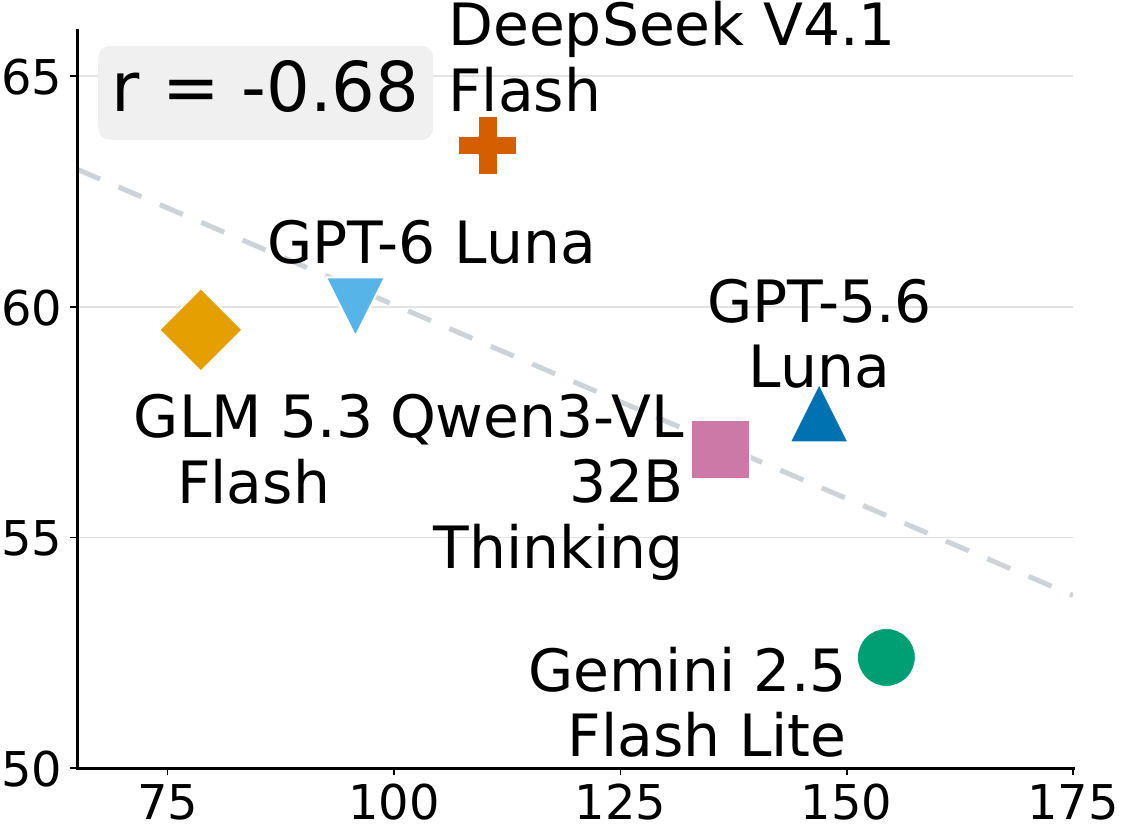}
\par\small c) tokens/query (k)
\end{minipage}%
\hfill
\begin{minipage}[t]{0.24\linewidth}
\centering
\includegraphics[width=\linewidth]{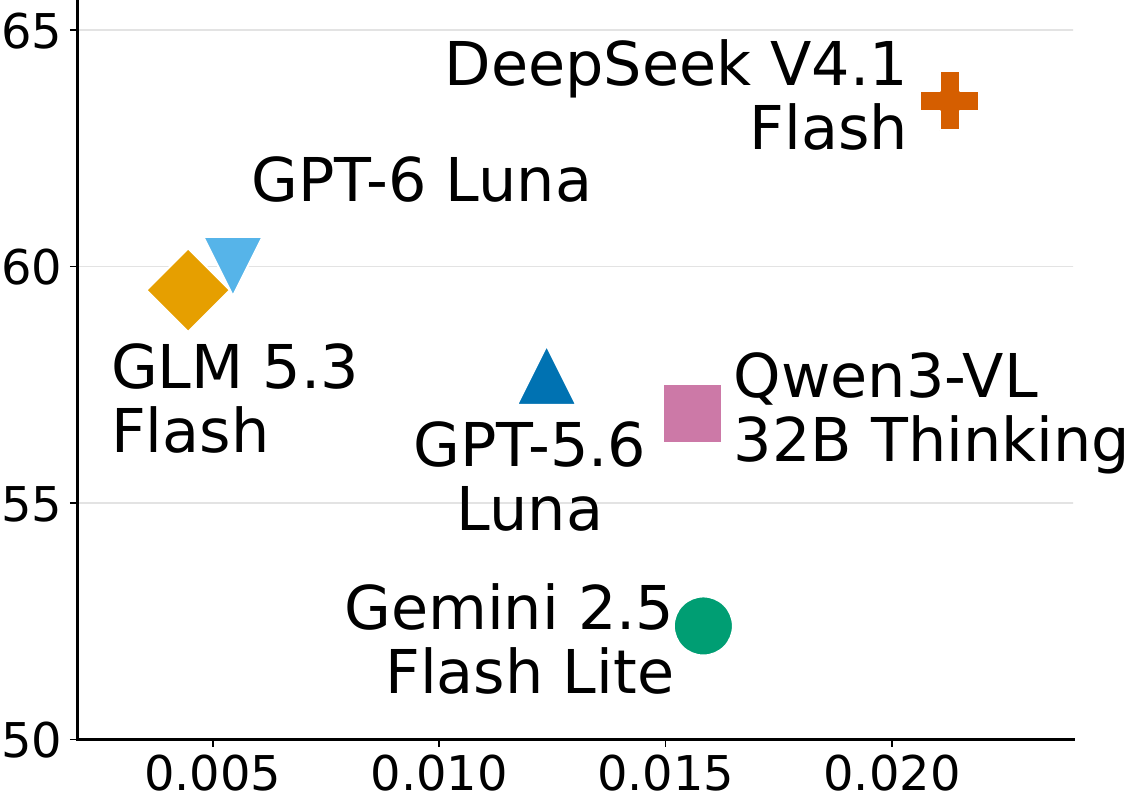}
\par\small d) cost/query (\$)
\end{minipage}%
\caption{
\textbf{\sname's performance \emph{v.s.} VLM capability and inference cost.} Each panel plots overall \stask gIoU against \textbf{a)} \emph{MMMU-Pro} score, which measures multimodal understanding and reasoning, \textbf{b)} \emph{Humanity's Last Exam} (HLE) score, measuring expert-level academic knowledge and reasoning, \textbf{c)} average token usage per query, and \textbf{d)} average VLM API cost per query. Dashed lines indicate linear fits, with Pearson's coefficient $r$ reported where shown. Benchmark scores are from Artificial Analysis~\cite{artificialanalysis2026}; GLM 5.3 Flash's MMMU-Pro score is missing from the source.
}
\label{fig:vlm_capability_cost}
\end{figure}

%% file: floats/tab_ablation_instruction.tex
\begin{figure}[t]
\centering\small
\begin{minipage}[t]{0.39\linewidth}
\newcommand{\close}{\hspace{3pt}}
\vspace{0pt}
\centering\small
\captionof{table}{\textbf{Detailed long-form queries improve \sname but overwhelm \samthreeagent.} Results are reported on the total split of \stask. \Quotes{short/long} denote the original class labels and the long queries used in \stask. Long queries improve \sname by 5.9\% in gIoU, but reduce \samthreeagent by 22.1\% in gIoU.}
\label{table:abinstruct}
\begin{tabular}{@{}l@{\close}|@{\close}l@{\close}|@{\close}c@{\close}|@{\close}c@{\close}|@{\close}c@{}}
\toprule
Method & Text & \metrics\\
\midrule
\samthreeagent & short        & \sota{46.4} & \sota{48.1} & 48.0 \\
\samthreeagent & long & 24.3        & 34.5        & \sota{30.7} \\
\midrule
\textbf{\ours} & short        & 51.0 & 51.3 & 41.7 \\
\textbf{\ours} & long & \sota{56.9} & \sota{66.5} & \sota{22.9} \\
\bottomrule
\end{tabular}
\end{minipage}
\hfill
\begin{minipage}[t]{0.59\linewidth}
\vspace{0pt}
\centering\small
\newcommand{\squareimage}[1]{\includegraphics[width=0.24\textwidth, height=0.24\textwidth]{#1}}
\newcommand{\close}{\hspace{1pt}}
\resizebox{\textwidth}{!}{
\begin{tabular}{@{}c@{\close}c@{\close}c@{\close}c@{}}
\ssamthreeagent short & \ssamthreeagent long & \sname short & \sname long \\
\squareimage{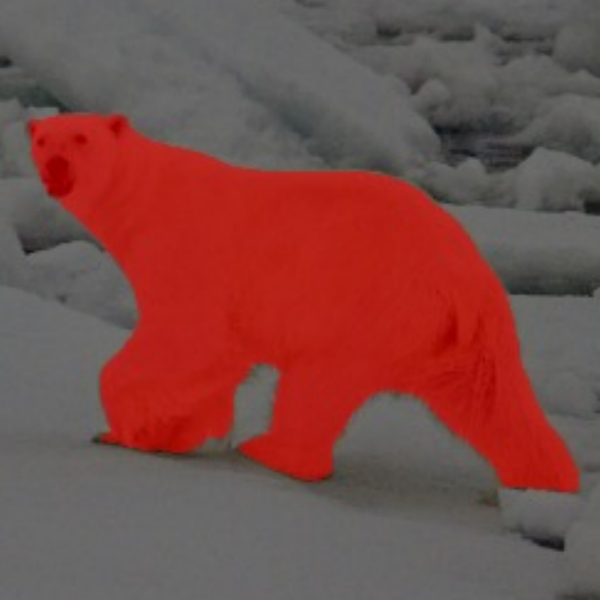} &
\squareimage{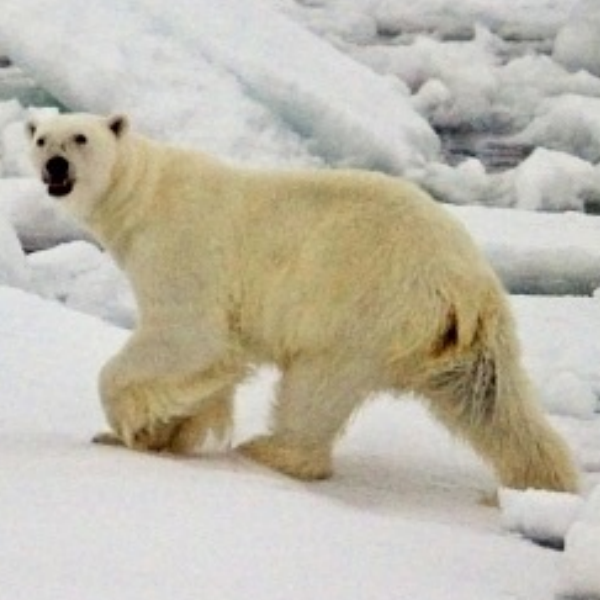} &
\squareimage{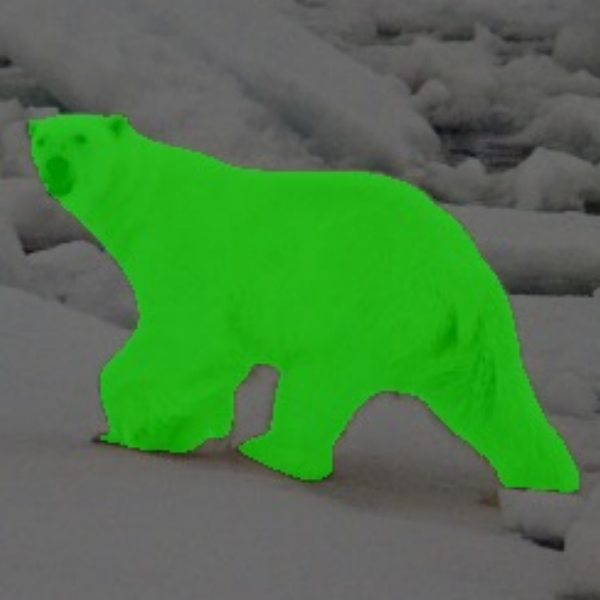} &
\squareimage{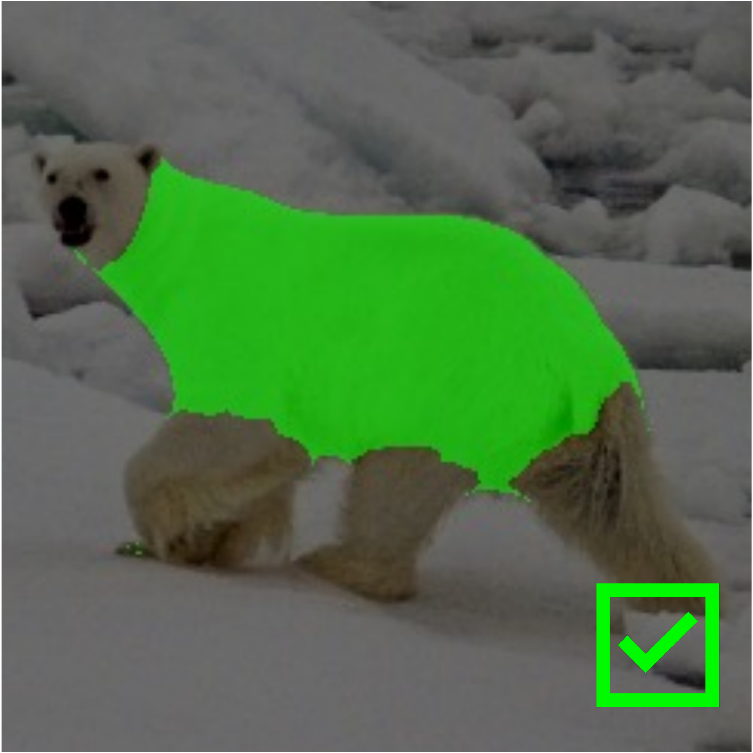}
\end{tabular}
}
\captionof{figure}{\textbf{Effect of detailed long-form queries.} Short query: \Quotes{\textit{the body of polar bear}}. Long query (shortened): \Quotes{\textit{Segment the torso of the polar bear, starting from below the head to just above the legs.}}. The short query is ambiguous, leading both methods to segment the entire bear. The long-form query resolves the ambiguity, but \samthreeagent (S.A.) completely fails. In contrast, \sname successfully follows the query and isolates the requested torso region.}
\label{fig:abinstructexample}
\end{minipage}
\end{figure}

%% file: floats/fig_reason_vs_iou.tex
\begin{figure}[h]
\centering\small
\begin{minipage}[t]{0.4\linewidth}
\vspace{0pt}
\centering
\includegraphics[width=\linewidth]{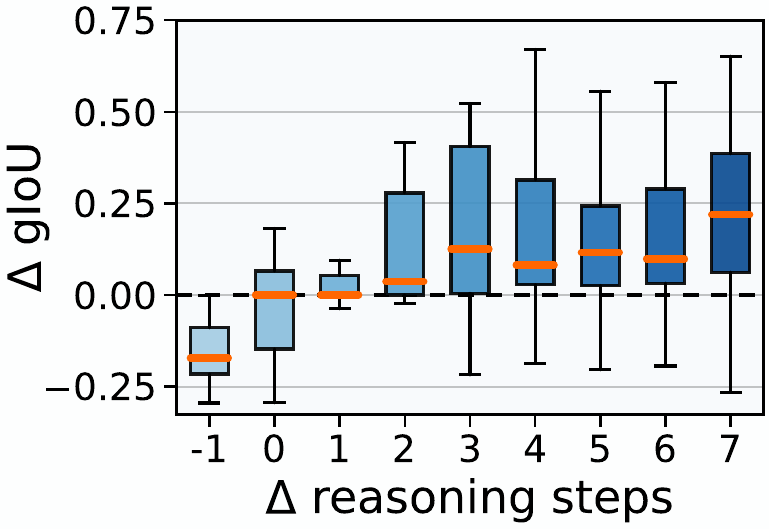}
\end{minipage}\hfill
\begin{minipage}[t]{0.58\linewidth}
\vspace{1pt}
\caption{\textbf{Additional reasoning steps correlate with improved segmentation quality.} The x-axis shows the number of more reasoning steps used by \sname than \samthreeagent, while the y-axis shows the corresponding gIoU improvement. Darker blue colors indicate bins containing more instances. The positive correlations suggest that additional long-horizon reasoning and refinement are often beneficial for challenging open ad-hoc concepts. \sname naturally has more reasoning steps than \samthreeagent due to the iterative visual construction.}
\label{fig:reasonvsiou}
\end{minipage}

\medskip
\begin{minipage}[t]{0.41\linewidth}
\vspace{0pt}
\centering\small
\setlength{\tabcolsep}{4pt}
\begin{tabular}{@{}lcccr@{}}
\toprule
Method & \shortstack{\samthree\\calls} & \shortstack{Reason.\\rounds} & \shortstack{Time per\\image}\\
\midrule
\samthreeagent & 3.1 & 3.8 & 5.2 s\\
\textbf{\sname} & 3.4 & 7.9 & 9.3 s\\
\bottomrule
\end{tabular}
\end{minipage}\hfill
\begin{minipage}[t]{0.57\linewidth}
\captionof{table}{\textbf{The additional reasoning rounds of \sname are effectively used for visual inspection and refinement.} We report average inference statistics on \stask with Qwen3-VL 32B Thinking hosted on an 8$\times$ NVIDIA A40 server under the same 20-round budget.}
\label{tab:runtime-main}
\end{minipage}
\end{figure}

%% file: floats/fig_qualitative.tex
\begin{figure*}[t]
\small\centering
\newcommand{\squareimage}[1]{\includegraphics[width=0.158\linewidth, height=0.158\linewidth]{#1}}
\newcommand{\colspace}{\hspace{8pt}}
\newcommand{\colspaceclose}{\hspace{1pt}}
\newcommand{\colspacenone}{\hspace{0pt}}
\newcommand{\parstablerow}[1]{
\squareimage{figures/fig_qualitative/#1_original} &
\squareimage{figures/fig_qualitative/#1_sam3agent_part} &
\squareimage{figures/fig_qualitative/#1_ours_instruct} &
\squareimage{figures/fig_qualitative/#1_gt}
}
\newcommand{\refcocomtablerow}[1]{
\squareimage{figures/fig_qualitative/ann#1_original} &
\squareimage{figures/fig_qualitative/ann#1_sam3agent} &
\squareimage{figures/fig_qualitative/ann#1_ours} &
\squareimage{figures/fig_qualitative/ann#1_gt}
}
\resizebox{\textwidth}{!}{
\renewcommand{\arraystretch}{0.5}
\begin{tabular}
{@{}c@{\colspaceclose}c@{\colspaceclose}c@{\colspaceclose}c@{\colspace}
c@{\colspaceclose}c@{\colspaceclose}c@{\colspaceclose}c@{}}
Image & \samthreeagent & \ours & GT & Image & \samthreeagent & \ours & GT \\ [2pt]
\parstablerow{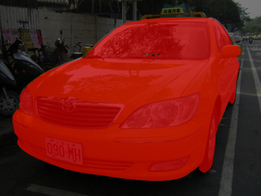} & \refcocomtablerow{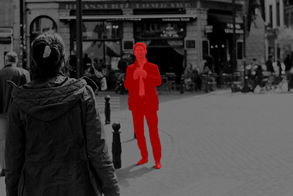} \\
\multicolumn{4}{l}{\textit{Segment the main structure of taxi, but no tires or side mirrors}} & \multicolumn{4}{l}{\textit{Legs of the man facing us in the middle}} \\ [4pt]
\parstablerow{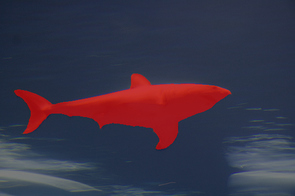} & \refcocomtablerow{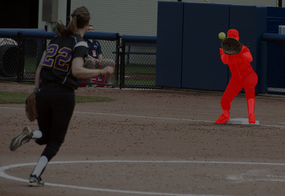} \\
\multicolumn{4}{l}{\textit{Segment the shark from where the head ends to just before the tail fin}} & \multicolumn{4}{l}{\textit{Torso of the guy on the right}} \\ [2pt]
\parstablerow{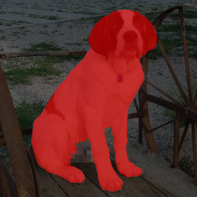} & \refcocomtablerow{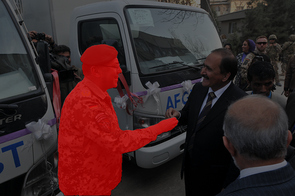} \\
\multicolumn{4}{l}{\textit{Segment the dog from below the neck to above the legs}} & \multicolumn{4}{l}{\textit{Green beret arm}}
\end{tabular}
}
\caption{\textbf{Qualitative comparisons between \samthreeagent and \sname on \stask (left) and \refcocom (right).} \samthreeagent often oversegments semantically related regions or confuses nearby concepts, while \sname precisely follows the text query through visual reasoning and iterative mask refinement. \stask prompts are shortened for clarity, while \refcocom prompts are original.}
\label{fig:qualitative}
\end{figure*}

%% file: sections/6_summary.tex
\textbf{Summary.} We present \sname, the first vision harnessing agent for open ad-hoc segmentation, and \stask, a new benchmark with detailed queries defining open ad-hoc concepts. \sname features a complete harness that coordinates strategy, constraints, scrutiny, and recovery around a persistent working mask, enabling the agent to inspect and refine its visual progress. We demonstrate \sname on \stask across six different VLMs, as well as on related standard benchmarks like \refcocom and \reasonseg. We conduct ablation studies and detailed analysis to understand where \sname's gain comes from. Our work points toward AI agents that go beyond invoking tools, using structured harnesses to maintain visual state, guide reasoning, inspect progress, and recover from errors.

%% file: sections/X1_appendix.tex
\appendix

\onecolumn
\begin{center}{\bf \Large \papertitle}\end{center}
\begin{center}{\Large Appendix}\end{center}

\hypersetup{linkcolor=black}
\etocdepthtag.toc{mtappendix}
\etocsettagdepth{mtchapter}{none}
\etocsettagdepth{mtappendix}{subsection}
\tableofcontents
\clearpage
\hypersetup{linkcolor=red}

\input{sections/X2_related}

\clearpage

\section{\sname Inference Algorithm}
\label{appendix:vasa}

Algorithm~\ref{alg:vasa} describes the interaction between the VLM, segmentation model, and harness. Each controller turn selects one tool call; it need not contain a segmentation call or a mask edit. Planning, candidate generation, optional individual-mask inspection, and editing can therefore occupy separate turns. The working mask changes only through an explicit edit.

\input{floats/algo_vasainference}

\clearpage
\input{sections/X3_implementation}

\section{\stask Construction Details}
\label{appendix:pars}

\textbf{Images and annotations.}
We construct \stask from the validation split of \partin~\cite{he2022partimagenet}, retaining its images and human-annotated part masks. Within each image, masks belonging to the same part category are merged by union. We focus on the \emph{body} category, whose boundaries vary across objects and are often underspecified by the class name alone.

\textbf{Detailed queries.}
For each object--part concept, we provide Qwen3-VL 8B Instruct with five randomly sampled examples from the corresponding class, each pairing the original image with its annotated target. The model generates a description specifying the regions to include, neighboring parts to exclude, and cues for locating their boundaries. The prompt asks for a definition that applies across examples, without references to annotation overlays or bounding boxes. The authors manually verify and correct each description, which is then reused for the corresponding concept across evaluation images. The demonstration images overlap the evaluation set, but the annotated demonstrations and ground-truth masks are not provided to the segmentation agent. The VLM generates only the textual definitions; all target masks remain the original human annotations. The short-query ablation replaces these descriptions with the original class names while keeping images and target masks fixed.

\textbf{Common and ad-hoc concepts.}
We also provide Qwen3-VL 8B Instruct with images overlaid with their ground-truth masks and ask it to classify the concepts as common or open ad-hoc based on familiarity. This yields 1,084 common and 937 ad-hoc concepts.

\clearpage
\section{Additional Qualitative Results}
\label{appendix:qualitative}

We provide additional qualitative comparisons among \lisa~\cite{lai2023lisa}, \segzero~\cite{liu2025segzero}, \samthreeagent~\cite{carion2026sam3}, and \ours on \stask (Figs.~\ref{suppl:fig-qualitative-pars1} and \ref{suppl:fig-qualitative-pars2}) and \refcocom~\cite{wang2024mres} (Figs.~\ref{suppl:fig-qualitative-refcocom1} and \ref{suppl:fig-qualitative-refcocom2}). The qualitative results show that baselines often produce overly broad masks or confuse the target with nearby related regions. In contrast, \sname more precisely isolates the requested regions across both \stask and \refcocom, suggesting that its visual construction workflow generalizes beyond open ad-hoc segmentation to fine-grained referring segmentation.

\input{floats/fig_qualitative_appendix}

%% file: sections/X2_related.tex
\section{Extended Related Work}
\label{appendix:related}

\textbf{Referring segmentation} focuses on segmenting image regions by explicit text descriptions. Most prior works either derive segmentation as a byproduct of image–text alignment models~\cite{radford2021clip, cherti2023openclip, zhai2023siglip, tschannen2025siglip2, jia2021align}, leveraging attention maps~\cite{woo2026caft, zhou2022extractfreedenselabels, LI2025111409, wang2024sclip, xiao2025flair, park2026freegrained, xu2025nepa} or patch-level similarity~\cite{zeng2025maskclippp, xu2022groupvit, ding2023maskclip, mukhoti2022pacl, ranasinghe2023perceptualgroup, yi2023simseg, Luo2023SegCLIP, ghiasi2022scaling, ni2023refdiff}, or directly train models with region–text alignment supervision~\cite{kuo2022findit, li2022lseg, yuan2024ovsam, li2024omgseg, xu2023san, gu2022vild, kamath2021mdetr, jiang2024trex2, li2023desco, li2022glip, zhang2022glipv2, liang2023ovseg, minderer2022simdet, minderer2024scalingovdet, liu2023groundingdino, ren2024groundingdino15, ren2024dinox, ren2024grounded, shen2024ape, xu2024mmquery, heng2025rodmllm, zhang2024evfsam, zou2022xdecoder, zou2023seem, xu2023odise, wang2024samclip, carion2026sam3}. Open ad-hoc segmentation generalizes it from grounding existing visual entities toward constructing arbitrary user-specified visual concepts through composition and exclusion reasoning. \gres and the \grefcoco dataset~\cite{liu2023gres} consider a special case of \oas, where the target concept is composed of multiple commonly seen objects.

\textbf{Referring part segmentation} focuses on segmenting existing part concepts. \semanticsam~\cite{li2023semanticsam} and \resany~\cite{wang2025resanything} leverages \sam~\cite{kirillov2023sam, ravi2024sam2} to generate part-level masks for training and inference. Other works~\cite{liu2023universalseg, wang2024mres, jang2025mmr, wei2023ovparts, peize2023vlpart, wan2025instructpart} have also explored closed-set part datasets like the \partin~\cite{he2022partimagenet} and introduced new part referring segmentation datasets like the \refcocom~\cite{wang2024mres}, along with others~\cite{chen2014pascalparts, ramanathan2023paco, wang2025resanything, jang2025mmr, wei2023ovparts}. Like with objects, these methods mainly cover the common parts. Since there are far more parts than objects, open ad-hoc segmentation becomes even more critical in the part domain.

\textbf{Reasoning segmentation} focuses on segmenting typically well-defined targets from \emph{implicit} text queries through commonsense and spatial reasoning. Most existing works use VLMs for reasoning and segmenters for mask decoding. Three types of reasoning paradigms prevail. \textbf{1)} Implicit reasoning, where the VLM predicts the target segment representation, keeping the reasoning process inaccessible~\cite{qian2026reasoningattend, lai2023lisa, yang2024lisapp, xia2024gsva, hanoona2023GLaMM, zhang2024omgllava, wei2024hyperseg, wang2024llmseg, chen2024sam4mllm, zhang2025psalm, wang2026xsam, li2026sam3i}, such as the \segtoken token in \lisa~\cite{lai2023lisa}. \textbf{2)} Text-based visual reasoning, where the VLM is tuned through RL~\cite{shao2024grpo} to reason in text~\cite{huang2026samr, yun2026star, liu2026visionreasoner, liu2025segzero, wang2026pixelthink, ren2023pixellm, zhu2025lens, hegde2026gensegr1}, such as the target's coordinates prediction in \segzero~\cite{liu2025segzero}. \textbf{3)} Grounded visual chain-of-thought, where each reasoning step is visually grounded, and subsequent steps build on both prior reasoning and grounding results~\cite{lu2025rsvp, dong2026cotreferring, lu2026coprs, wu2025groundedcot, man2025argus, shao2024visualcot, wang2024segllm}, such as SegLLM~\cite{wang2024segllm}. Open ad-hoc segmentation is an orthogonal direction, reasoning about what constitutes a visual concept rather than how to find the existing concept hinted by the user.

%% file: floats/algo_vasainference.tex
\begin{algorithm}[h!]
\caption{\sname controller loop (one tool call per turn)}
\label{alg:vasa}
\KwIn{Image $I$, query $q$, VLM $A$, segmenter \samthree, threshold $T$}
\KwOut{Constructed mask, or fallback prediction}
Initialize working mask $M\leftarrow\emptyset$, candidates $\mathcal C\leftarrow\emptyset$, strategy $\pi$, history $h$, warning $w$, and counter $n\leftarrow0$\;
\While{the interaction is active}{
    Assemble context from $I,q,M,\mathcal C,\pi,h,w$ and system instructions\;
    Request one controller response from $A$; increment $n$ and record the response\;
    Parse the tool call and apply the supported execution checks\;
    \If{a formatting or validation error is detected}{
        Set corrective warning $w$; \textbf{continue}\;
    }
    \uIf{\texttt{set\_strategy}}{
        Store the supplied plan as $\pi$\;
    }
    \uElseIf{\texttt{segment\_phrase}}{
        Generate new candidates $\mathcal C$ from the supplied phrase; keep $M$ unchanged\;
        \If{no candidates are returned}{Set warning $w$; \textbf{continue}\;}
        Render numbered candidate masks\;
    }
    \uElseIf{\texttt{examine\_each\_mask}}{
        Inspect candidates individually with auxiliary VLM calls\;
        Retain accepted candidates and return an overlay or warning; keep $M$ unchanged\;
    }
    \uElseIf{\texttt{update\_working\_mask}}{
        Form $C$ as the union of selected valid candidates\;
        Apply $M\leftarrow M\cup C$, $M\leftarrow M\setminus C$, or $M\leftarrow C$\;
        Mark $M$ as updated and render its overlay\;
    }
    \ElseIf{\texttt{return\_final\_output} or \texttt{report\_no\_mask}}{
        \textbf{break}\;
    }
    \If{$n>T$}{\textbf{break}\;}
}
\Return{$M$ if explicitly updated; otherwise latest candidates, or an empty prediction}\;
\end{algorithm}

%% file: sections/X3_implementation.tex
\section{Implementation Details}
\label{appendix:implementation}

\textbf{Models and inference settings.}
We use pretrained \samthree~\cite{carion2026sam3} as the segmentation tool, without task-specific fine-tuning. We use Qwen3-VL 32B Thinking~\cite{bai2025qwen3vl} for the ablations and additional benchmarks, hosting it locally on eight A40 GPUs. The other five VLMs are accessed through OpenRouter. Images sent to the VLM are downscaled by a factor of two in each dimension, while mask operations retain the original image resolution. We use the serving endpoints' default decoding settings, except for GLM 5.3 Flash, whose reasoning effort is set to \texttt{low} to avoid exhausting the generation budget before producing a tool call. Per-response token limits vary by model.

\textbf{Interaction budget.}
Both \sname and \samthreeagent use a 20-round budget. A round may involve planning, inspection, or mask editing and need not include a segmentation call. Auxiliary per-mask inspection calls and error-recovery turns count toward the budget; retries caused by network failures or API rate limits do not.

\textbf{Visual state and feedback.}
At each turn, the controller receives the original image and query, recorded strategy, current working mask, latest candidates, and a compact history of tool calls. Earlier intermediate images and free-form responses are omitted. New segmentation candidates do not overwrite the working mask, which changes only through deterministic binary edits. Per-mask inspection checks candidates against the latest segmentation phrase using the original image, a mask overlay, and a zoomed view. Execution errors produce feedback while preserving the working mask. The controller makes all semantic judgments from the query and visual observations, without access to ground-truth masks.

%% file: floats/fig_qualitative_appendix.tex
\newcommand{\squareimage}[1]{\includegraphics[width=0.158\linewidth, height=0.158\linewidth]{#1}}
\newcommand{\colspace}{\hspace{8pt}}
\newcommand{\colspaceclose}{\hspace{1pt}}
\newcommand{\colspacenone}{\hspace{0pt}}
\newcommand{\tablerow}[2]{
\squareimage{figures/fig_qualitative_appendix/#1/ann#2_original} &
\squareimage{figures/fig_qualitative_appendix/#1/ann#2_lisa} &
\squareimage{figures/fig_qualitative_appendix/#1/ann#2_segzero} &
\squareimage{figures/fig_qualitative_appendix/#1/ann#2_sam3agent} &
\squareimage{figures/fig_qualitative_appendix/#1/ann#2_ours} &
\squareimage{figures/fig_qualitative_appendix/#1/ann#2_gt} \\ [-1pt]
\multicolumn{6}{l}{\includegraphics[width=0.98\linewidth]{figures/fig_qualitative_appendix/#1/ann#2_instruction.pdf}}
}

\begin{figure*}[h!]
\small\centering
\resizebox{\textwidth}{!}{
\renewcommand{\arraystretch}{0.5}
\begin{tabular}
{@{}c@{\colspace}c@{\colspaceclose}c@{\colspaceclose}c@{\colspace}c@{\colspace}c@{}}
Image & \lisa & \segzero & \samthreeagent & \ours & GT \\ [2pt]
\tablerow{pars1}{524} \\
\tablerow{pars1}{2486} \\
\tablerow{pars1}{4970} \\
\tablerow{pars1}{6274} \\
\tablerow{pars1}{6330} \\
\tablerow{pars1}{6402}
\end{tabular}
}
\caption{\textbf{Additional qualitative results of \sname and baselines on \stask.}}
\label{suppl:fig-qualitative-pars1}
\end{figure*}

\begin{figure*}[h!]
\small\centering
\resizebox{\textwidth}{!}{
\renewcommand{\arraystretch}{0.5}
\begin{tabular}
{@{}c@{\colspace}c@{\colspaceclose}c@{\colspaceclose}c@{\colspace}c@{\colspace}c@{}}
Image & \lisa & \segzero & \samthreeagent & \ours & GT \\ [2pt]
\tablerow{pars2}{3897} \\
\tablerow{pars2}{4023} \\
\tablerow{pars2}{1646} \\
\tablerow{pars2}{62} \\
\tablerow{pars2}{1994} \\
\tablerow{pars2}{2189} \\
\tablerow{pars2}{8923}
\end{tabular}
}
\caption{\textbf{Additional qualitative results of \sname and baselines on \stask} (continued).}
\label{suppl:fig-qualitative-pars2}
\end{figure*}

\begin{figure*}[h!]
\small\centering
\resizebox{\textwidth}{!}{
\renewcommand{\arraystretch}{0.5}
\begin{tabular}
{@{}c@{\colspace}c@{\colspaceclose}c@{\colspaceclose}c@{\colspace}c@{\colspace}c@{}}
Image & \lisa & \segzero & \samthreeagent & \ours & GT \\ [2pt]
\tablerow{refcocom}{22} \\
\tablerow{refcocom}{67} \\
\tablerow{refcocom}{75} \\
\tablerow{refcocom}{82} \\
\tablerow{refcocom}{88} \\
\tablerow{refcocom}{230} \\
\tablerow{refcocom}{108} \\
\tablerow{refcocom}{203}
\end{tabular}
}
\caption{\textbf{Additional results of \sname and baselines on \refcocom.}}
\label{suppl:fig-qualitative-refcocom1}
\end{figure*}

\begin{figure*}[h!]
\small\centering
\resizebox{\textwidth}{!}{
\renewcommand{\arraystretch}{0.5}
\begin{tabular}
{@{}c@{\colspace}c@{\colspaceclose}c@{\colspaceclose}c@{\colspace}c@{\colspace}c@{}}
Image & \lisa & \segzero & \samthreeagent & \ours & GT \\ [2pt]
\tablerow{refcocom}{30} \\
\tablerow{refcocom}{41} \\
\tablerow{refcocom}{49} \\
\tablerow{refcocom}{52} \\
\tablerow{refcocom}{248} \\
\tablerow{refcocom}{213} \\
\tablerow{refcocom}{277} \\
\tablerow{refcocom}{202} \\
\end{tabular}
}
\caption{\textbf{Additional results of \sname and baselines on \refcocom} (continued).}
\label{suppl:fig-qualitative-refcocom2}
\end{figure*}

%% file: main.bbl
\begin{thebibliography}{111}
\providecommand{\natexlab}[1]{#1}
\providecommand{\url}[1]{\texttt{#1}}
\expandafter\ifx\csname urlstyle\endcsname\relax
  \providecommand{\doi}[1]{doi: #1}\else
  \providecommand{\doi}{doi: \begingroup \urlstyle{rm}\Url}\fi

\bibitem[{Artificial Analysis}(2026)]{artificialanalysis2026}
{Artificial Analysis}.
\newblock Model evaluations.
\newblock \url{https://artificialanalysis.ai/evaluations}, 2026.

\bibitem[Bai et~al.(2025)]{bai2025qwen3vl}
Shuai Bai et~al.
\newblock Qwen3-vl technical report, 2025.
\newblock URL \url{https://arxiv.org/abs/2511.21631}.

\bibitem[Bao et~al.(2024)Bao, Sun, Ma, Zheng, Guo, Zhao, Zheng, and
  Wang]{bao2024cores}
Xiaoyi Bao, Siyang Sun, Shuailei Ma, Kecheng Zheng, Yuxin Guo, Guosheng Zhao,
  Yun Zheng, and Xingang Wang.
\newblock Cores: Orchestrating the dance of reasoning and segmentation.
\newblock In \emph{European Conference on Computer Vision}, pp.\  187--204.
  Springer, 2024.

\bibitem[Carion et~al.(2026)Carion, Gustafson, Hu, Debnath, Hu, Coll-Vinent,
  Ryali, Alwala, Khedr, Huang, Lei, Ma, Guo, Kalla, Marks, Greer, Wang, Sun,
  R{\"a}dle, Afouras, Mavroudi, Xu, Wu, Zhou, Momeni, HAZRA, Ding, Vaze,
  Porcher, Li, Li, Kamath, Cheng, Dollar, Ravi, Saenko, Zhang, and
  Feichtenhofer]{carion2026sam3}
Nicolas Carion, Laura Gustafson, Yuan-Ting Hu, Shoubhik Debnath, Ronghang Hu,
  Didac~Suris Coll-Vinent, Chaitanya Ryali, Kalyan~Vasudev Alwala, Haitham
  Khedr, Andrew Huang, Jie Lei, Tengyu Ma, Baishan Guo, Arpit Kalla, Markus
  Marks, Joseph Greer, Meng Wang, Peize Sun, Roman R{\"a}dle, Triantafyllos
  Afouras, Effrosyni Mavroudi, Katherine Xu, Tsung-Han Wu, Yu~Zhou, Liliane
  Momeni, RISHI HAZRA, Shuangrui Ding, Sagar Vaze, Francois Porcher, Feng Li,
  Siyuan Li, Aishwarya Kamath, Ho~Kei Cheng, Piotr Dollar, Nikhila Ravi, Kate
  Saenko, Pengchuan Zhang, and Christoph Feichtenhofer.
\newblock {SAM} 3: Segment anything with concepts.
\newblock In \emph{The Fourteenth International Conference on Learning
  Representations}, 2026.
\newblock URL \url{https://openreview.net/forum?id=r35clVtGzw}.

\bibitem[Chen et~al.(2014)Chen, Mottaghi, Liu, Fidler, Urtasun, and
  Yuille]{chen2014pascalparts}
Xianjie Chen, Roozbeh Mottaghi, Xiaobai Liu, Sanja Fidler, Raquel Urtasun, and
  Alan Yuille.
\newblock Detect what you can: Detecting and representing objects using
  holistic models and body parts, 2014.
\newblock URL \url{https://arxiv.org/abs/1406.2031}.

\bibitem[Chen et~al.(2024)Chen, Li, Sun, Wang, and Chen]{chen2024sam4mllm}
Yi-Chia Chen, Wei-Hua Li, Cheng Sun, Yu-Chiang~Frank Wang, and Chu-Song Chen.
\newblock Sam4mllm: Enhance multi-modal large language model for referring
  expression segmentation, 2024.
\newblock URL \url{https://arxiv.org/abs/2409.10542}.

\bibitem[Cherti et~al.(2023)Cherti, Beaumont, Wightman, Wortsman, Ilharco,
  Gordon, Schuhmann, Schmidt, and Jitsev]{cherti2023openclip}
Mehdi Cherti, Romain Beaumont, Ross Wightman, Mitchell Wortsman, Gabriel
  Ilharco, Cade Gordon, Christoph Schuhmann, Ludwig Schmidt, and Jenia Jitsev.
\newblock Reproducible scaling laws for contrastive language-image learning.
\newblock In \emph{Proceedings of the IEEE/CVF Conference on Computer Vision
  and Pattern Recognition}, pp.\  2818--2829, 2023.

\bibitem[Comanici et~al.(2025)]{comanici2025gemini25}
Gheorghe Comanici et~al.
\newblock Gemini 2.5: Pushing the frontier with advanced reasoning,
  multimodality, long context, and next generation agentic capabilities, 2025.
\newblock URL \url{https://arxiv.org/abs/2507.06261}.

\bibitem[DeepSeek-AI et~al.(2026)]{deepseekai2026deepseekv41flash}
DeepSeek-AI et~al.
\newblock Deepseek-v4.1-flash: Pushing the limits of kv cache compression,
  2026.
\newblock URL \url{https://arxiv.org/abs/2609.19969}.

\bibitem[Ding et~al.(2023)Ding, Wang, and Tu]{ding2023maskclip}
Zheng Ding, Jieke Wang, and Zhuowen Tu.
\newblock Open-vocabulary universal image segmentation with maskclip.
\newblock In \emph{International Conference on Machine Learning}, 2023.

\bibitem[Dong et~al.(2026)Dong, Figueroa, Zhao, Kafle, Kuen, Ding, Cohen, and
  Fu]{dong2026cotreferring}
Qihua Dong, Luis Figueroa, Handong Zhao, Kushal Kafle, Jason Kuen, Zhihong
  Ding, Scott Cohen, and Yun Fu.
\newblock Cot referring: Improving referring expression tasks with grounded
  reasoning, 2026.
\newblock URL \url{https://arxiv.org/abs/2510.06243}.

\bibitem[Du et~al.(2026)Du, Li, Fan, Zhang, Pan, and Zhang]{du2026samveteran}
Tianyuan Du, Haopeng Li, Zhen Fan, Jiarui Zhang, Panwang Pan, and Yang Zhang.
\newblock {SAM}-veteran: An {MLLM}-based human-like {SAM} agent for reasoning
  segmentation.
\newblock In \emph{The Fourteenth International Conference on Learning
  Representations}, 2026.
\newblock URL \url{https://openreview.net/forum?id=oN55r8iJJW}.

\bibitem[Ghiasi et~al.(2022)Ghiasi, Gu, Cui, and Lin]{ghiasi2022scaling}
Golnaz Ghiasi, Xiuye Gu, Yin Cui, and Tsung-Yi Lin.
\newblock Scaling open-vocabulary image segmentation with image-level labels,
  2022.
\newblock URL \url{https://arxiv.org/abs/2112.12143}.

\bibitem[GLM-5-Team et~al.(2026)]{glm5team2026glm5vibecodingagentic}
GLM-5-Team et~al.
\newblock Glm-5: from vibe coding to agentic engineering, 2026.
\newblock URL \url{https://arxiv.org/abs/2602.15763}.

\bibitem[Gu et~al.(2022)Gu, Lin, Kuo, and Cui]{gu2022vild}
Xiuye Gu, Tsung-Yi Lin, Weicheng Kuo, and Yin Cui.
\newblock Open-vocabulary object detection via vision and language knowledge
  distillation.
\newblock In \emph{International Conference on Learning Representations}, 2022.
\newblock URL \url{https://openreview.net/forum?id=lL3lnMbR4WU}.

\bibitem[Han et~al.(2026)Han, Hu, Qiu, Wu, and He]{vista2026}
Qiushi Han, Keya Hu, Linlu Qiu, Cathy Wu, and Kaiming He.
\newblock {VISTA}: A visual harness for reasoning in an interactive world, aug
  2026.
\newblock URL \url{https://vista-research.github.io/}.

\bibitem[Hashimoto(2026)]{hashimoto2026aiadoption}
Mitchell Hashimoto.
\newblock My ai adoption journey.
\newblock \url{https://mitchellh.com/writing/my-ai-adoption-journey}, 2026.

\bibitem[He et~al.(2022)He, Yang, Yang, Kortylewski, Yuan, Chen, Liu, Yang, Yu,
  and Yuille]{he2022partimagenet}
Ju~He, Shuo Yang, Shaokang Yang, Adam Kortylewski, Xiaoding Yuan, Jie-Neng
  Chen, Shuai Liu, Cheng Yang, Qihang Yu, and Alan Yuille.
\newblock Partimagenet: A large, high-quality dataset of parts, 2022.
\newblock URL \url{https://arxiv.org/abs/2112.00933}.

\bibitem[He et~al.(2025)He, Zhang, Gao, Li, Hong, Chen, Jiang, Fu, and
  Zhang]{he2025rsagent}
Xingqi He, Yujie Zhang, Shuyong Gao, Wenjie Li, Lingyi Hong, Mingxi Chen,
  Kaixun Jiang, Jiyuan Fu, and Wenqiang Zhang.
\newblock Rsagent: Learning to reason and act for text-guided segmentation via
  multi-turn tool invocations, 2025.
\newblock URL \url{https://arxiv.org/abs/2512.24023}.

\bibitem[Hegde et~al.(2026)Hegde, Chacko, Banerjee, and
  Mahesh]{hegde2026gensegr1}
Sandesh Hegde, Jaison~Saji Chacko, Debarshi Banerjee, and Uma Mahesh.
\newblock Genseg-r1: Rl-driven vision-language grounding for fine-grained
  referring segmentation, 2026.
\newblock URL \url{https://arxiv.org/abs/2602.09701}.

\bibitem[Huang et~al.(2026)Huang, Xu, Zhou, Liu, Xiao, Ou, Ji, Li, and
  Yuan]{huang2026samr}
Jiaqi Huang, Zunnan Xu, Jun Zhou, Ting Liu, Yicheng Xiao, Mingwen Ou, Bowen Ji,
  Xiu Li, and Kehong Yuan.
\newblock Sam-r1: Leveraging sam for reward feedback in multimodal segmentation
  via reinforcement learning, 2026.
\newblock URL \url{https://arxiv.org/abs/2505.22596}.

\bibitem[Jang et~al.(2025)Jang, Cho, Lee, Kim, and Kim]{jang2025mmr}
Donggon Jang, Yucheol Cho, Suin Lee, Taehyeon Kim, and Dae-Shik Kim.
\newblock Mmr: A large-scale benchmark dataset for multi-target and
  multi-granularity reasoning segmentation, 2025.
\newblock URL \url{https://arxiv.org/abs/2503.13881}.

\bibitem[Jia et~al.(2021)Jia, Yang, Xia, Chen, Parekh, Pham, Le, Sung, Li, and
  Duerig]{jia2021align}
Chao Jia, Yinfei Yang, Ye~Xia, Yi-Ting Chen, Zarana Parekh, Hieu Pham, Quoc~V.
  Le, Yunhsuan Sung, Zhen Li, and Tom Duerig.
\newblock Scaling up visual and vision-language representation learning with
  noisy text supervision, 2021.
\newblock URL \url{https://arxiv.org/abs/2102.05918}.

\bibitem[Jiang et~al.(2024)Jiang, Li, Zeng, Ren, Liu, and
  Zhang]{jiang2024trex2}
Qing Jiang, Feng Li, Zhaoyang Zeng, Tianhe Ren, Shilong Liu, and Lei Zhang.
\newblock T-rex2: Towards generic object detection via text-visual prompt
  synergy, 2024.

\bibitem[Kamath et~al.(2021)Kamath, Singh, LeCun, Misra, Synnaeve, and
  Carion]{kamath2021mdetr}
Aishwarya Kamath, Mannat Singh, Yann LeCun, Ishan Misra, Gabriel Synnaeve, and
  Nicolas Carion.
\newblock Mdetr--modulated detection for end-to-end multi-modal understanding.
\newblock \emph{arXiv preprint arXiv:2104.12763}, 2021.

\bibitem[Kirillov et~al.(2023)Kirillov, Mintun, Ravi, Mao, Rolland, Gustafson,
  Xiao, Whitehead, Berg, Lo, Doll{\'a}r, and Girshick]{kirillov2023sam}
Alexander Kirillov, Eric Mintun, Nikhila Ravi, Hanzi Mao, Chloe Rolland, Laura
  Gustafson, Tete Xiao, Spencer Whitehead, Alexander~C. Berg, Wan-Yen Lo, Piotr
  Doll{\'a}r, and Ross Girshick.
\newblock Segment anything.
\newblock \emph{arXiv:2304.02643}, 2023.

\bibitem[Kuo et~al.(2022)Kuo, Bertsch, Li, Piergiovanni, Saffar, and
  Angelova]{kuo2022findit}
Weicheng Kuo, Fred Bertsch, Wei Li, AJ~Piergiovanni, Mohammad Saffar, and
  Anelia Angelova.
\newblock Findit: Generalized localization with natural language queries, 2022.
\newblock URL \url{https://arxiv.org/abs/2203.17273}.

\bibitem[Lai et~al.(2023)Lai, Tian, Chen, Li, Yuan, Liu, and Jia]{lai2023lisa}
Xin Lai, Zhuotao Tian, Yukang Chen, Yanwei Li, Yuhui Yuan, Shu Liu, and Jiaya
  Jia.
\newblock Lisa: Reasoning segmentation via large language model.
\newblock \emph{arXiv preprint arXiv:2308.00692}, 2023.

\bibitem[Li et~al.(2022)Li, Weinberger, Belongie, Koltun, and
  Ranftl]{li2022lseg}
Boyi Li, Kilian~Q Weinberger, Serge Belongie, Vladlen Koltun, and Rene Ranftl.
\newblock Language-driven semantic segmentation.
\newblock In \emph{International Conference on Learning Representations}, 2022.
\newblock URL \url{https://openreview.net/forum?id=RriDjddCLN}.

\bibitem[Li et~al.(2023{\natexlab{a}})Li, Zhang, Sun, Zou, Liu, Yang, Li,
  Zhang, and Gao]{li2023semanticsam}
Feng Li, Hao Zhang, Peize Sun, Xueyan Zou, Shilong Liu, Jianwei Yang, Chunyuan
  Li, Lei Zhang, and Jianfeng Gao.
\newblock Semantic-sam: Segment and recognize anything at any granularity.
\newblock \emph{arXiv preprint arXiv:2307.04767}, 2023{\natexlab{a}}.

\bibitem[Li et~al.(2026{\natexlab{a}})Li, Feng, Guo, Huang, Ji, Bi, Piao,
  Zhang, Zhao, Chen, Zou, Lu, and Cheng]{li2026sam3i}
Jingjing Li, Yue Feng, Yuchen Guo, Jincai Huang, Wei Ji, Qi~Bi, Yongri Piao,
  Miao Zhang, Xiaoqi Zhao, Qiang Chen, Shihao Zou, Huchuan Lu, and Li~Cheng.
\newblock Sam3-i: Segment anything with instructions, 2026{\natexlab{a}}.
\newblock URL \url{https://arxiv.org/abs/2512.04585}.

\bibitem[Li* et~al.(2022)Li*, Zhang*, Zhang*, Yang, Li, Zhong, Wang, Yuan,
  Zhang, Hwang, Chang, and Gao]{li2022glip}
Liunian~Harold Li*, Pengchuan Zhang*, Haotian Zhang*, Jianwei Yang, Chunyuan
  Li, Yiwu Zhong, Lijuan Wang, Lu~Yuan, Lei Zhang, Jenq-Neng Hwang, Kai-Wei
  Chang, and Jianfeng Gao.
\newblock Grounded language-image pre-training.
\newblock In \emph{CVPR}, 2022.

\bibitem[Li et~al.(2023{\natexlab{b}})Li, Dou, Peng, and Chang]{li2023desco}
Liunian~Harold Li, Zi-Yi Dou, Nanyun Peng, and Kai-Wei Chang.
\newblock Desco: Learning object recognition with rich language descriptions.
\newblock In \emph{Thirty-seventh Conference on Neural Information Processing
  Systems}, 2023{\natexlab{b}}.
\newblock URL \url{https://openreview.net/forum?id=J2Cso0wWZX}.

\bibitem[Li et~al.(2024)Li, Yuan, Li, Ding, Wu, Zhang, Li, Chen, and
  Loy]{li2024omgseg}
Xiangtai Li, Haobo Yuan, Wei Li, Henghui Ding, Size Wu, Wenwei Zhang, Yining
  Li, Kai Chen, and Chen~Change Loy.
\newblock Omg-seg: Is one model good enough for all segmentation?, 2024.
\newblock URL \url{https://arxiv.org/abs/2401.10229}.

\bibitem[Li et~al.(2025)Li, Wang, Duan, Zhang, and Li]{LI2025111409}
Yi~Li, Hualiang Wang, Yiqun Duan, Jiheng Zhang, and Xiaomeng Li.
\newblock A closer look at the explainability of contrastive language-image
  pre-training.
\newblock \emph{Pattern Recognition}, 162:\penalty0 111409, 2025.
\newblock ISSN 0031-3203.
\newblock \doi{https://doi.org/10.1016/j.patcog.2025.111409}.
\newblock URL
  \url{https://www.sciencedirect.com/science/article/pii/S003132032500069X}.

\bibitem[Li et~al.(2026{\natexlab{b}})Li, Jin, Liu, Liu, Prasad, D'Eramo, and
  Chalvatzaki]{li2026harbor}
Zechu Li, Yufeng Jin, Xiaoyang Liu, Puze Liu, Vignesh Prasad, Carlo D'Eramo,
  and Georgia Chalvatzaki.
\newblock Harbor: A harness framework for agentic robot reinforcement learning,
  2026{\natexlab{b}}.
\newblock URL \url{https://arxiv.org/abs/2606.08610}.

\bibitem[Liang et~al.(2023)Liang, Wu, Dai, Li, Zhao, Zhang, Zhang, Vajda, and
  Marculescu]{liang2023ovseg}
Feng Liang, Bichen Wu, Xiaoliang Dai, Kunpeng Li, Yinan Zhao, Hang Zhang,
  Peizhao Zhang, Peter Vajda, and Diana Marculescu.
\newblock Open-vocabulary semantic segmentation with mask-adapted clip.
\newblock In \emph{Proceedings of the IEEE/CVF Conference on Computer Vision
  and Pattern Recognition}, pp.\  7061--7070, 2023.

\bibitem[Liu et~al.(2023{\natexlab{a}})Liu, Ding, and Jiang]{liu2023gres}
Chang Liu, Henghui Ding, and Xudong Jiang.
\newblock Gres: Generalized referring expression segmentation,
  2023{\natexlab{a}}.
\newblock URL \url{https://arxiv.org/abs/2306.00968}.

\bibitem[Liu et~al.(2023{\natexlab{b}})Liu, Zeng, Ren, Li, Zhang, Yang, Li,
  Yang, Su, Zhu, et~al.]{liu2023groundingdino}
Shilong Liu, Zhaoyang Zeng, Tianhe Ren, Feng Li, Hao Zhang, Jie Yang, Chunyuan
  Li, Jianwei Yang, Hang Su, Jun Zhu, et~al.
\newblock Grounding dino: Marrying dino with grounded pre-training for open-set
  object detection.
\newblock \emph{arXiv preprint arXiv:2303.05499}, 2023{\natexlab{b}}.

\bibitem[Liu et~al.(2023{\natexlab{c}})Liu, Zhang, Wang, Wang, Yang, and
  Tang]{liu2023universalseg}
Yong Liu, Cairong Zhang, Yitong Wang, Jiahao Wang, Yujiu Yang, and Yansong
  Tang.
\newblock Universal segmentation at arbitrary granularity with language
  instruction.
\newblock \emph{arXiv preprint arXiv:2312.01623}, 2023{\natexlab{c}}.

\bibitem[Liu et~al.(2025)Liu, Peng, Zhong, Yue, Lu, Yu, and
  Jia]{liu2025segzero}
Yuqi Liu, Bohao Peng, Zhisheng Zhong, Zihao Yue, Fanbin Lu, Bei Yu, and Jiaya
  Jia.
\newblock Seg-zero: Reasoning-chain guided segmentation via cognitive
  reinforcement.
\newblock \emph{arXiv preprint arXiv:2503.06520}, 2025.

\bibitem[Liu et~al.(2026)Liu, Qu, Zhong, Peng, Liu, Yu, and
  Jia]{liu2026visionreasoner}
Yuqi Liu, Tianyuan Qu, Zhisheng Zhong, Bohao Peng, Shu Liu, Bei Yu, and Jiaya
  Jia.
\newblock Visionreasoner: Unified reasoning-integrated visual perception via
  reinforcement learning.
\newblock In \emph{The Fourteenth International Conference on Learning
  Representations}, 2026.

\bibitem[Lu et~al.(2025)Lu, Cao, Wu, Li, Tang, Ji, Wu, Wu, and Zhu]{lu2025rsvp}
Yi~Lu, Jiawang Cao, Yongliang Wu, Bozheng Li, Licheng Tang, Yangguang Ji, Chong
  Wu, Jay Wu, and Wenbo Zhu.
\newblock Rsvp: Reasoning segmentation via visual prompting and multi-modal
  chain-of-thought, 2025.
\newblock URL \url{https://arxiv.org/abs/2506.04277}.

\bibitem[Lu et~al.(2026)Lu, Li, Wang, Feng, Chen, Chen, and Wang]{lu2026coprs}
Zhenyu Lu, Liupeng Li, Jinpeng Wang, Yan Feng, Bin Chen, Ke~Chen, and Yaowei
  Wang.
\newblock Co{PRS}: Learning positional prior from chain-of-thought for
  reasoning segmentation.
\newblock In \emph{The Fourteenth International Conference on Learning
  Representations}, 2026.
\newblock URL \url{https://openreview.net/forum?id=Fcsop01h40}.

\bibitem[Luo et~al.(2023)Luo, Bao, Wu, He, and Li]{Luo2023SegCLIP}
Huaishao Luo, Junwei Bao, Youzheng Wu, Xiaodong He, and Tianrui Li.
\newblock {SegCLIP}: Patch aggregation with learnable centers for
  open-vocabulary semantic segmentation.
\newblock \emph{ICML}, 2023.

\bibitem[Man et~al.(2025)Man, Huang, Liu, Sheng, Liu, Gui, Kautz, Wang, and
  Yu]{man2025argus}
Yunze Man, De-An Huang, Guilin Liu, Shiwei Sheng, Shilong Liu, Liang-Yan Gui,
  Jan Kautz, Yu-Xiong Wang, and Zhiding Yu.
\newblock Argus: Vision-centric reasoning with grounded chain-of-thought, 2025.
\newblock URL \url{https://arxiv.org/abs/2505.23766}.

\bibitem[Minderer et~al.(2022)Minderer, Gritsenko, Stone, Neumann, Weissenborn,
  Dosovitskiy, Mahendran, Arnab, Dehghani, Shen, Wang, Zhai, Kipf, and
  Houlsby]{minderer2022simdet}
Matthias Minderer, Alexey Gritsenko, Austin Stone, Maxim Neumann, Dirk
  Weissenborn, Alexey Dosovitskiy, Aravindh Mahendran, Anurag Arnab, Mostafa
  Dehghani, Zhuoran Shen, Xiao Wang, Xiaohua Zhai, Thomas Kipf, and Neil
  Houlsby.
\newblock Simple open-vocabulary object detection with vision transformers,
  2022.
\newblock URL \url{https://arxiv.org/abs/2205.06230}.

\bibitem[Minderer et~al.(2024)Minderer, Gritsenko, and
  Houlsby]{minderer2024scalingovdet}
Matthias Minderer, Alexey Gritsenko, and Neil Houlsby.
\newblock Scaling open-vocabulary object detection, 2024.
\newblock URL \url{https://arxiv.org/abs/2306.09683}.

\bibitem[Mukhoti et~al.(2022)Mukhoti, Lin, Poursaeed, Wang, Shah, Torr, and
  Lim]{mukhoti2022pacl}
Jishnu Mukhoti, Tsung-Yu Lin, Omid Poursaeed, Rui Wang, Ashish Shah, Philip
  H.~S. Torr, and Ser-Nam Lim.
\newblock Open vocabulary semantic segmentation with patch aligned contrastive
  learning, 2022.
\newblock URL \url{https://arxiv.org/abs/2212.04994}.

\bibitem[Ni et~al.(2023)Ni, Zhang, Feng, Li, Guo, and Zuo]{ni2023refdiff}
Minheng Ni, Yabo Zhang, Kailai Feng, Xiaoming Li, Yiwen Guo, and Wangmeng Zuo.
\newblock Ref-diff: Zero-shot referring image segmentation with generative
  models.
\newblock \emph{arXiv preprint arXiv:2308.16777}, 2023.

\bibitem[{OpenAI}(2026{\natexlab{a}})]{openai2026gpt56}
{OpenAI}.
\newblock {GPT-5.6}: Frontier intelligence that scales with your ambition.
\newblock \url{https://openai.com/index/gpt-5-6/}, 2026{\natexlab{a}}.

\bibitem[{OpenAI}(2026{\natexlab{b}})]{openai2026gpt6}
{OpenAI}.
\newblock Introducing {GPT-6 Sol and Luna}.
\newblock \url{https://openai.com/index/introducing-gpt-6-sol-and-luna/},
  2026{\natexlab{b}}.

\bibitem[{OpenAI}(2026{\natexlab{c}})]{openai2026harness}
{OpenAI}.
\newblock Harness engineering.
\newblock \url{https://openai.com/index/harness-engineering/},
  2026{\natexlab{c}}.

\bibitem[Park et~al.(2026)Park, Wang, and Yu]{park2026freegrained}
Seulki Park, Zilin Wang, and Stella~X. Yu.
\newblock Free-grained hierarchical visual recognition, 2026.
\newblock URL \url{https://arxiv.org/abs/2510.14737}.

\bibitem[Qian et~al.(2026)Qian, Yin, and Dou]{qian2026reasoningattend}
Rui Qian, Xin Yin, and Dejing Dou.
\newblock Reasoning to attend: Try to understand how <seg> token works, 2026.
\newblock URL \url{https://arxiv.org/abs/2412.17741}.

\bibitem[Radford et~al.(2021)Radford, Kim, Hallacy, Ramesh, Goh, Agarwal,
  Sastry, Askell, Mishkin, Clark, Krueger, and Sutskever]{radford2021clip}
Alec Radford, Jong~Wook Kim, Chris Hallacy, Aditya Ramesh, Gabriel Goh,
  Sandhini Agarwal, Girish Sastry, Amanda Askell, Pamela Mishkin, Jack Clark,
  Gretchen Krueger, and Ilya Sutskever.
\newblock Learning transferable visual models from natural language
  supervision, 2021.
\newblock URL \url{https://arxiv.org/abs/2103.00020}.

\bibitem[Ramanathan et~al.(2023)Ramanathan, Kalia, Petrovic, Wen, Zheng, Guo,
  Wang, Marquez, Kovvuri, Kadian, Mousavi, Song, Dubey, and
  Mahajan]{ramanathan2023paco}
Vignesh Ramanathan, Anmol Kalia, Vladan Petrovic, Yi~Wen, Baixue Zheng, Baishan
  Guo, Rui Wang, Aaron Marquez, Rama Kovvuri, Abhishek Kadian, Amir Mousavi,
  Yiwen Song, Abhimanyu Dubey, and Dhruv Mahajan.
\newblock {PACO}: Parts and attributes of common objects.
\newblock In \emph{arXiv preprint arXiv:2301.01795}, 2023.

\bibitem[Ranasinghe et~al.(2023)Ranasinghe, McKinzie, Ravi, Yang, Toshev, and
  Shlens]{ranasinghe2023perceptualgroup}
Kanchana Ranasinghe, Brandon McKinzie, Sachin Ravi, Yinfei Yang, Alexander
  Toshev, and Jonathon Shlens.
\newblock Perceptual grouping in contrastive vision-language models, 2023.
\newblock URL \url{https://arxiv.org/abs/2210.09996}.

\bibitem[Rasheed et~al.(2024)Rasheed, Maaz, Shaji, Shaker, Khan, Cholakkal,
  Anwer, Xing, Yang, and Khan]{hanoona2023GLaMM}
Hanoona Rasheed, Muhammad Maaz, Sahal Shaji, Abdelrahman Shaker, Salman Khan,
  Hisham Cholakkal, Rao~M. Anwer, Eric Xing, Ming-Hsuan Yang, and Fahad~S.
  Khan.
\newblock Glamm: Pixel grounding large multimodal model.
\newblock \emph{The IEEE/CVF Conference on Computer Vision and Pattern
  Recognition}, 2024.

\bibitem[Ravi et~al.(2024)Ravi, Gabeur, Hu, Hu, Ryali, Ma, Khedr, R{\"a}dle,
  Rolland, Gustafson, Mintun, Pan, Alwala, Carion, Wu, Girshick, Doll{\'a}r,
  and Feichtenhofer]{ravi2024sam2}
Nikhila Ravi, Valentin Gabeur, Yuan-Ting Hu, Ronghang Hu, Chaitanya Ryali,
  Tengyu Ma, Haitham Khedr, Roman R{\"a}dle, Chloe Rolland, Laura Gustafson,
  Eric Mintun, Junting Pan, Kalyan~Vasudev Alwala, Nicolas Carion, Chao-Yuan
  Wu, Ross Girshick, Piotr Doll{\'a}r, and Christoph Feichtenhofer.
\newblock Sam 2: Segment anything in images and videos.
\newblock \emph{arXiv preprint arXiv:2408.00714}, 2024.
\newblock URL \url{https://arxiv.org/abs/2408.00714}.

\bibitem[Ren et~al.(2024{\natexlab{a}})Ren, Chen, Jiang, Zeng, Xiong, Liu, Ma,
  Shen, Gao, Jiang, Chen, Song, Zhang, Huang, Gao, Liu, Zhang, Li, Yu, and
  Zhang]{ren2024dinox}
Tianhe Ren, Yihao Chen, Qing Jiang, Zhaoyang Zeng, Yuda Xiong, Wenlong Liu,
  Zhengyu Ma, Junyi Shen, Yuan Gao, Xiaoke Jiang, Xingyu Chen, Zhuheng Song,
  Yuhong Zhang, Hongjie Huang, Han Gao, Shilong Liu, Hao Zhang, Feng Li, Kent
  Yu, and Lei Zhang.
\newblock Dino-x: A unified vision model for open-world object detection and
  understanding, 2024{\natexlab{a}}.
\newblock URL \url{https://arxiv.org/abs/2411.14347}.

\bibitem[Ren et~al.(2024{\natexlab{b}})Ren, Jiang, Liu, Zeng, Liu, Gao, Huang,
  Ma, Jiang, Chen, Xiong, Zhang, Li, Tang, Yu, and
  Zhang]{ren2024groundingdino15}
Tianhe Ren, Qing Jiang, Shilong Liu, Zhaoyang Zeng, Wenlong Liu, Han Gao,
  Hongjie Huang, Zhengyu Ma, Xiaoke Jiang, Yihao Chen, Yuda Xiong, Hao Zhang,
  Feng Li, Peijun Tang, Kent Yu, and Lei Zhang.
\newblock Grounding dino 1.5: Advance the "edge" of open-set object detection,
  2024{\natexlab{b}}.
\newblock URL \url{https://arxiv.org/abs/2405.10300}.

\bibitem[Ren et~al.(2024{\natexlab{c}})Ren, Liu, Zeng, Lin, Li, Cao, Chen,
  Huang, Chen, Yan, Zeng, Zhang, Li, Yang, Li, Jiang, and
  Zhang]{ren2024grounded}
Tianhe Ren, Shilong Liu, Ailing Zeng, Jing Lin, Kunchang Li, He~Cao, Jiayu
  Chen, Xinyu Huang, Yukang Chen, Feng Yan, Zhaoyang Zeng, Hao Zhang, Feng Li,
  Jie Yang, Hongyang Li, Qing Jiang, and Lei Zhang.
\newblock Grounded sam: Assembling open-world models for diverse visual tasks,
  2024{\natexlab{c}}.

\bibitem[Ren et~al.(2023)Ren, Huang, Wei, Zhao, Fu, Feng, and
  Jin]{ren2023pixellm}
Zhongwei Ren, Zhicheng Huang, Yunchao Wei, Yao Zhao, Dongmei Fu, Jiashi Feng,
  and Xiaojie Jin.
\newblock Pixellm: Pixel reasoning with large multimodal model.
\newblock \emph{arXiv preprint arXiv:2312.02228}, 2023.

\bibitem[Shao et~al.(2024{\natexlab{a}})Shao, Qian, Xiao, Song, Zong, Wang,
  Liu, and Li]{shao2024visualcot}
Hao Shao, Shengju Qian, Han Xiao, Guanglu Song, Zhuofan Zong, Letian Wang,
  Yu~Liu, and Hongsheng Li.
\newblock Visual cot: Unleashing chain-of-thought reasoning in multi-modal
  language models, 2024{\natexlab{a}}.

\bibitem[Shao et~al.(2024{\natexlab{b}})Shao, Wang, Zhu, Xu, Song, Bi, Zhang,
  Zhang, Li, Wu, and Guo]{shao2024grpo}
Zhihong Shao, Peiyi Wang, Qihao Zhu, Runxin Xu, Junxiao Song, Xiao Bi, Haowei
  Zhang, Mingchuan Zhang, Y.~K. Li, Y.~Wu, and Daya Guo.
\newblock Deepseekmath: Pushing the limits of mathematical reasoning in open
  language models, 2024{\natexlab{b}}.
\newblock URL \url{https://arxiv.org/abs/2402.03300}.

\bibitem[Shen et~al.(2024)Shen, Fu, Chen, Zhang, Li, Sun, Wu, Lin, and
  Ji]{shen2024ape}
Yunhang Shen, Chaoyou Fu, Peixian Chen, Mengdan Zhang, Ke~Li, Xing Sun,
  Yunsheng Wu, Shaohui Lin, and Rongrong Ji.
\newblock Aligning and prompting everything all at once for universal visual
  perception.
\newblock In \emph{Proceedings of the IEEE/CVF Conference on Computer Vision
  and Pattern Recognition}, 2024.

\bibitem[Sun et~al.(2023)Sun, Chen, Zhu, Xiao, Luo, Xie, and
  Yan]{peize2023vlpart}
Peize Sun, Shoufa Chen, Chenchen Zhu, Fanyi Xiao, Ping Luo, Saining Xie, and
  Zhicheng Yan.
\newblock Going denser with open-vocabulary part segmentation.
\newblock \emph{arXiv preprint arXiv:2305.11173}, 2023.

\bibitem[Tschannen et~al.(2025)Tschannen, Gritsenko, Wang, Naeem,
  Alabdulmohsin, Parthasarathy, Evans, Beyer, Xia, Mustafa, H\'enaff, Harmsen,
  Steiner, and Zhai]{tschannen2025siglip2}
Michael Tschannen, Alexey Gritsenko, Xiao Wang, Muhammad~Ferjad Naeem, Ibrahim
  Alabdulmohsin, Nikhil Parthasarathy, Talfan Evans, Lucas Beyer, Ye~Xia, Basil
  Mustafa, Olivier H\'enaff, Jeremiah Harmsen, Andreas Steiner, and Xiaohua
  Zhai.
\newblock Siglip 2: Multilingual vision-language encoders with improved
  semantic understanding, localization, and dense features.
\newblock \emph{arXiv preprint arXiv:2502.14786}, 2025.

\bibitem[Vats et~al.(2025)Vats, Rathee, and Davis]{vats2025guidelineseg}
Vanshika Vats, Ashwani Rathee, and James Davis.
\newblock Guideline-consistent segmentation via multi-agent refinement, 2025.
\newblock URL \url{https://arxiv.org/abs/2509.04687}.

\bibitem[Wan et~al.(2025)Wan, Xie, Zhang, Lin, Wang, Stepputtis, Ramanan, and
  Sycara]{wan2025instructpart}
Zifu Wan, Yaqi Xie, Ce~Zhang, Zhiqiu Lin, Zihan Wang, Simon Stepputtis, Deva
  Ramanan, and Katia~P. Sycara.
\newblock {I}nstruct{P}art: Task-oriented part segmentation with instruction
  reasoning.
\newblock In \emph{Proceedings of the 63rd Annual Meeting of the Association
  for Computational Linguistics (Volume 1: Long Papers)}, pp.\  24202--24227,
  Vienna, Austria, July 2025. Association for Computational Linguistics.
\newblock ISBN 979-8-89176-251-0.
\newblock \doi{10.18653/v1/2025.acl-long.1179}.
\newblock URL \url{https://aclanthology.org/2025.acl-long.1179/}.

\bibitem[Wang et~al.(2024{\natexlab{a}})Wang, Mei, and Yuille]{wang2024sclip}
Feng Wang, Jieru Mei, and Alan Yuille.
\newblock Sclip: Rethinking self-attention for dense vision-language inference,
  2024{\natexlab{a}}.
\newblock URL \url{https://arxiv.org/abs/2312.01597}.

\bibitem[Wang et~al.(2026{\natexlab{a}})Wang, Qiao, Jie, Huang, Feng, Zheng,
  Ma, Lan, and Liang]{wang2026xsam}
Hao Wang, Limeng Qiao, Zequn Jie, Zhijian Huang, Chengjian Feng, Qingfang
  Zheng, Lin Ma, Xiangyuan Lan, and Xiaodan Liang.
\newblock X-sam: From segment anything to any segmentation, 2026{\natexlab{a}}.
\newblock URL \url{https://arxiv.org/abs/2508.04655}.

\bibitem[Wang et~al.(2024{\natexlab{b}})Wang, Vasu, Faghri, Vemulapalli,
  Farajtabar, Mehta, Rastegari, Tuzel, and Pouransari]{wang2024samclip}
Haoxiang Wang, Pavan Kumar~Anasosalu Vasu, Fartash Faghri, Raviteja
  Vemulapalli, Mehrdad Farajtabar, Sachin Mehta, Mohammad Rastegari, Oncel
  Tuzel, and Hadi Pouransari.
\newblock Sam-clip: Merging vision foundation models towards semantic and
  spatial understanding, 2024{\natexlab{b}}.
\newblock URL \url{https://arxiv.org/abs/2310.15308}.

\bibitem[Wang \& Ke(2024)Wang and Ke]{wang2024llmseg}
Junchi Wang and Lei Ke.
\newblock Llm-seg: Bridging image segmentation and large language model
  reasoning.
\newblock \emph{arXiv preprint arXiv:2404.08767}, 2024.

\bibitem[Wang et~al.(2026{\natexlab{b}})Wang, Wang, Li, Ban, Chen, Ge, Qin, Li,
  and Zhu]{wang2026thea}
Qi~Wang, Tianyi Wang, Chengyang Li, Shikun Ban, Yurun Chen, Yizhong Ge, Jason
  Qin, Chengtai Li, and Wentao Zhu.
\newblock Towards the harness of embodied agents, 2026{\natexlab{b}}.
\newblock URL \url{https://arxiv.org/abs/2608.11246}.

\bibitem[Wang \& Zhang(2025)Wang and Zhang]{wang2025resanything}
Ruiqi Wang and Hao Zhang.
\newblock Resanything: Attribute prompting for arbitrary referring
  segmentation, 2025.
\newblock URL \url{https://arxiv.org/abs/2505.02867}.

\bibitem[Wang et~al.(2026{\natexlab{c}})Wang, Fang, Kong, Li, Xu, Yang, Li,
  Zhu, and Wang]{wang2026pixelthink}
Song Wang, Gongfan Fang, Lingdong Kong, Xiangtai Li, Jianyun Xu, Sheng Yang,
  Qiang Li, Jianke Zhu, and Xinchao Wang.
\newblock Pixelthink: Towards efficient chain-of-pixel reasoning,
  2026{\natexlab{c}}.
\newblock URL \url{https://openreview.net/forum?id=wvxq3qNzHB}.

\bibitem[Wang et~al.(2024{\natexlab{c}})Wang, Yue, Zhang, Guo, He, Wang, and
  Liu]{wang2024mres}
Wenxuan Wang, Tongtian Yue, Yisi Zhang, Longteng Guo, Xingjian He, Xinlong
  Wang, and Jing Liu.
\newblock Unveiling parts beyond objects:towards finer-granularity referring
  expression segmentation, 2024{\natexlab{c}}.
\newblock URL \url{https://arxiv.org/abs/2312.08007}.

\bibitem[Wang et~al.(2024{\natexlab{d}})Wang, Zhang, Li, Kallidromitis, Li,
  Kato, Kozuka, and Darrell]{wang2024segllm}
XuDong Wang, Shaolun Zhang, Shufan Li, Konstantinos Kallidromitis, Kehan Li,
  Yusuke Kato, Kazuki Kozuka, and Trevor Darrell.
\newblock Segllm: Multi-round reasoning segmentation.
\newblock \emph{arXiv preprint arXiv:2410.18923}, 2024{\natexlab{d}}.

\bibitem[Wang et~al.(2025)Wang, Mo, Yu, Behpour, and Ren]{wang2025open}
Zilin Wang, Sangwoo Mo, Stella~X. Yu, Sima Behpour, and Liu Ren.
\newblock Open ad-hoc categorization with contextualized feature learning.
\newblock In \emph{Proceedings of the IEEE/CVF Conference on Computer Vision
  and Pattern Recognition}, 2025.

\bibitem[Wei et~al.(2024)Wei, Zhong, Tan, Liu, Zhao, Hu, and
  Yang]{wei2024hyperseg}
Cong Wei, Yujie Zhong, Haoxian Tan, Yong Liu, Zheng Zhao, Jie Hu, and Yujiu
  Yang.
\newblock Hyperseg: Towards universal visual segmentation with large language
  model, 2024.
\newblock URL \url{https://arxiv.org/abs/2411.17606}.

\bibitem[Wei et~al.(2023)Wei, Yue, Zhang, Kong, Liu, and Pang]{wei2023ovparts}
Meng Wei, Xiaoyu Yue, Wenwei Zhang, Shu Kong, Xihui Liu, and Jiangmiao Pang.
\newblock Ov-parts: Towards open-vocabulary part segmentation.
\newblock In \emph{Thirty-seventh Conference on Neural Information Processing
  Systems Datasets and Benchmarks Track}, 2023.

\bibitem[Woo et~al.(2026)Woo, Wang, Pak, Mo, and Yu]{woo2026caft}
Byeongju Woo, Zilin Wang, Byeonghyun Pak, Sangwoo Mo, and Stella~X. Yu.
\newblock Aligning forest and trees in images and long captions for visually
  grounded understanding, 2026.
\newblock URL \url{https://arxiv.org/abs/2602.02977}.

\bibitem[Wu et~al.(2025)Wu, Yang, Zhou, Fang, Song, Sun, and
  Ji]{wu2025groundedcot}
Qiong Wu, Xiangcong Yang, Yiyi Zhou, Chenxin Fang, Baiyang Song, Xiaoshuai Sun,
  and Rongrong Ji.
\newblock Grounded chain-of-thought for multimodal large language models, 2025.
\newblock URL \url{https://arxiv.org/abs/2503.12799}.

\bibitem[Xia et~al.(2024)Xia, Han, Han, Pan, Song, and Huang]{xia2024gsva}
Zhuofan Xia, Dongchen Han, Yizeng Han, Xuran Pan, Shiji Song, and Gao Huang.
\newblock Gsva: Generalized segmentation via multimodal large language models.
\newblock In \emph{Proceedings of the IEEE/CVF Conference on Computer Vision
  and Pattern Recognition (CVPR)}, pp.\  3858--3869, June 2024.

\bibitem[Xiao et~al.(2025)Xiao, Kim, Georgescu, Akata, and
  Alaniz]{xiao2025flair}
Rui Xiao, Sanghwan Kim, Mariana-Iuliana Georgescu, Zeynep Akata, and Stephan
  Alaniz.
\newblock Flair: Vlm with fine-grained language-informed image representations.
\newblock In \emph{CVPR}, 2025.

\bibitem[Xiao et~al.(2026)Xiao, Xie, Zhang, Lin, Fu, Xue, Lu, Yang, Dai, Wang,
  Wu, Wang, Sastry, Goldberg, Fan, Zhu, and Shi]{xiao2026enpire}
Wenli Xiao, Jia Xie, Tonghe Zhang, Haotian Lin, Letian~"Max" Fu, Haoru Xue,
  Jalen Lu, Yi~Yang, Cunxi Dai, Zi~Wang, Jimmy Wu, Guanzhi Wang, S.~Shankar
  Sastry, Ken Goldberg, Linxi~"Jim" Fan, Yuke Zhu, and Guanya Shi.
\newblock Enpire: Agentic robot policy self-improvement in the real world,
  2026.
\newblock URL \url{https://arxiv.org/abs/2606.19980}.

\bibitem[Xu et~al.(2022)Xu, De~Mello, Liu, Byeon, Breuel, Kautz, and
  Wang]{xu2022groupvit}
Jiarui Xu, Shalini De~Mello, Sifei Liu, Wonmin Byeon, Thomas Breuel, Jan Kautz,
  and Xiaolong Wang.
\newblock Groupvit: Semantic segmentation emerges from text supervision.
\newblock \emph{arXiv preprint arXiv:2202.11094}, 2022.

\bibitem[Xu et~al.(2023{\natexlab{a}})Xu, Liu, Vahdat, Byeon, Wang, and
  Mello]{xu2023odise}
Jiarui Xu, Sifei Liu, Arash Vahdat, Wonmin Byeon, Xiaolong Wang, and Shalini~De
  Mello.
\newblock Open-vocabulary panoptic segmentation with text-to-image diffusion
  models, 2023{\natexlab{a}}.
\newblock URL \url{https://arxiv.org/abs/2303.04803}.

\bibitem[Xu et~al.(2023{\natexlab{b}})Xu, Zhang, Wei, Hu, and Bai]{xu2023san}
Mengde Xu, Zheng Zhang, Fangyun Wei, Han Hu, and Xiang Bai.
\newblock San: Side adapter network for open-vocabulary semantic segmentation.
\newblock \emph{IEEE Transactions on Pattern Analysis and Machine
  Intelligence}, 2023{\natexlab{b}}.

\bibitem[Xu et~al.(2025)Xu, Ma, Chai, Chen, Jin, Chai, Xie, and Yu]{xu2025nepa}
Sihan Xu, Ziqiao Ma, Wenhao Chai, Xuweiyi Chen, Weiyang Jin, Joyce Chai,
  Saining Xie, and Stella~X. Yu.
\newblock Next-embedding prediction makes strong vision learners, 2025.
\newblock URL \url{https://arxiv.org/abs/2512.16922}.

\bibitem[Xu et~al.(2024)Xu, Zhang, Fu, Chen, Yang, Li, and Xu]{xu2024mmquery}
Yifan Xu, Mengdan Zhang, Chaoyou Fu, Peixian Chen, Xiaoshan Yang, Ke~Li, and
  Changsheng Xu.
\newblock Multi-modal queried object detection in the wild.
\newblock \emph{Advances in Neural Information Processing Systems}, 36, 2024.

\bibitem[Yang et~al.(2024)Yang, Qu, Lai, Tian, Peng, Liu, and
  Jia]{yang2024lisapp}
Senqiao Yang, Tianyuan Qu, Xin Lai, Zhuotao Tian, Bohao Peng, Shu Liu, and
  Jiaya Jia.
\newblock Lisa++: An improved baseline for reasoning segmentation with large
  language model, 2024.
\newblock URL \url{https://arxiv.org/abs/2312.17240}.

\bibitem[Ye et~al.(2025)Ye, You, Lin, Ji, Dai, and Cao]{ye2025evolsam3}
Kai Ye, Xiaotong You, Jianghang Lin, Jiayi Ji, Pingyang Dai, and Liujuan Cao.
\newblock Evolving, not training: Zero-shot reasoning segmentation via
  evolutionary prompting, 2025.
\newblock URL \url{https://arxiv.org/abs/2512.24702}.

\bibitem[Yi et~al.(2023)Yi, Cui, Wu, Yang, Yoshie, and Lu]{yi2023simseg}
Muyang Yi, Quan Cui, Hao Wu, Cheng Yang, Osamu Yoshie, and Hongtao Lu.
\newblock A simple framework for text-supervised semantic segmentation.
\newblock In \emph{Proceedings of the IEEE/CVF Conference on Computer Vision
  and Pattern Recognition (CVPR)}, pp.\  7071--7080, 2023.

\bibitem[Yin et~al.(2025)Yin, Ren, Yan, Ding, and Hao]{heng2025rodmllm}
Heng Yin, Yuqiang Ren, Ke~Yan, Shouhong Ding, and Yongtao Hao.
\newblock Rod-mllm: Towards more reliable object detection in multimodal large
  language models.
\newblock In \emph{2025 IEEE/CVF Conference on Computer Vision and Pattern
  Recognition (CVPR)}, pp.\  14358--14368, 2025.
\newblock \doi{10.1109/CVPR52734.2025.01339}.

\bibitem[Yuan et~al.(2024)Yuan, Li, Zhou, Li, Chen, and Loy]{yuan2024ovsam}
Haobo Yuan, Xiangtai Li, Chong Zhou, Yining Li, Kai Chen, and Chen~Change Loy.
\newblock Open-vocabulary sam: Segment and recognize twenty-thousand classes
  interactively.
\newblock In \emph{ECCV}, 2024.

\bibitem[Yun et~al.(2026)Yun, Lee, Bae, Jun, Cho, and Ro]{yun2026star}
Seokju Yun, Dongheon Lee, Noori Bae, Jaesung Jun, Chanseul Cho, and Youngmin
  Ro.
\newblock Star: Segment anything reasoner, 2026.
\newblock URL \url{https://arxiv.org/abs/2603.14382}.

\bibitem[Zeng et~al.(2025)Zeng, Li, Zhou, Li, Hou, and
  Cheng]{zeng2025maskclippp}
Quan-Sheng Zeng, Yunheng Li, Daquan Zhou, Guanbin Li, Qibin Hou, and Ming-Ming
  Cheng.
\newblock High-quality mask tuning matters for open-vocabulary segmentation,
  2025.
\newblock URL \url{https://arxiv.org/abs/2412.11464}.

\bibitem[Zhai et~al.(2023)Zhai, Mustafa, Kolesnikov, and Beyer]{zhai2023siglip}
Xiaohua Zhai, Basil Mustafa, Alexander Kolesnikov, and Lucas Beyer.
\newblock Sigmoid loss for language image pre-training, 2023.
\newblock URL \url{https://arxiv.org/abs/2303.15343}.

\bibitem[Zhang et~al.(2022)Zhang, Zhang, Hu, Chen, Li, Dai, Wang, Yuan, Hwang,
  and Gao]{zhang2022glipv2}
Haotian* Zhang, Pengchuan* Zhang, Xiaowei Hu, Yen-Chun Chen, Liunian~Harold Li,
  Xiyang Dai, Lijuan Wang, Lu~Yuan, Jenq-Neng Hwang, and Jianfeng Gao.
\newblock Glipv2: Unifying localization and vision-language understanding.
\newblock \emph{arXiv preprint arXiv:2206.05836}, 2022.

\bibitem[Zhang et~al.(2024{\natexlab{a}})Zhang, Li, Fei, Yuan, Wu, Ji, Loy, and
  YAN]{zhang2024omgllava}
Tao Zhang, Xiangtai Li, Hao Fei, Haobo Yuan, Shengqiong Wu, Shunping Ji,
  Chen~Change Loy, and Shuicheng YAN.
\newblock {OMG}-{LL}a{VA}: Bridging image-level, object-level, pixel-level
  reasoning and understanding.
\newblock In \emph{The Thirty-eighth Annual Conference on Neural Information
  Processing Systems}, 2024{\natexlab{a}}.
\newblock URL \url{https://openreview.net/forum?id=WeoNd6PRqS}.

\bibitem[Zhang et~al.(2026)Zhang, Zhang, Gao, Li, Liu, Zhu, Qiu, Yan, Liu,
  Tang, Fang, Nie, Wei, Wang, Ding, and Yu]{zhang2026harnessvla}
Yixian Zhang, Huanming Zhang, Feng Gao, Xiao Li, Zhihao Liu, Chunyang Zhu,
  Jiaxing Qiu, Yuchen Yan, Jiyuan Liu, Wenhao Tang, Zhengru Fang, Yi~Nie,
  Changxu Wei, Yu~Wang, Wenbo Ding, and Chao Yu.
\newblock Harness vla: Steering frozen vlas into reliable manipulation
  primitives via memory-guided agents, 2026.
\newblock URL \url{https://arxiv.org/abs/2607.08448}.

\bibitem[Zhang et~al.(2024{\natexlab{b}})Zhang, Cheng, Hu, Liu, Liu, Ran, Chen,
  Liu, and Wang]{zhang2024evfsam}
Yuxuan Zhang, Tianheng Cheng, Rui Hu, Lei Liu, Heng Liu, Longjin Ran, Xiaoxin
  Chen, Wenyu Liu, and Xinggang Wang.
\newblock Evf-sam: Early vision-language fusion for text-prompted segment
  anything model.
\newblock \emph{arXiv preprint arXiv:2406.20076}, 2024{\natexlab{b}}.

\bibitem[Zhang et~al.(2025)Zhang, Ma, Zhang, and Bai]{zhang2025psalm}
Zheng Zhang, Yeyao Ma, Enming Zhang, and Xiang Bai.
\newblock Psalm: Pixelwise segmentation with large multi-modal model.
\newblock In \emph{European Conference on Computer Vision}, pp.\  74--91.
  Springer, 2025.

\bibitem[Zhou et~al.(2022)Zhou, Loy, and Dai]{zhou2022extractfreedenselabels}
Chong Zhou, Chen~Change Loy, and Bo~Dai.
\newblock Extract free dense labels from clip, 2022.
\newblock URL \url{https://arxiv.org/abs/2112.01071}.

\bibitem[Zhu et~al.(2025{\natexlab{a}})Zhu, Ouyang, Zhang, Cheng, Hu, Shen,
  Ran, Chen, Yu, Liu, and Wang]{zhu2025lens}
Lianghui Zhu, Bin Ouyang, Yuxuan Zhang, Tianheng Cheng, Rui Hu, Haocheng Shen,
  Longjin Ran, Xiaoxin Chen, Li~Yu, Wenyu Liu, and Xinggang Wang.
\newblock Lens: Learning to segment anything with unified reinforced reasoning,
  2025{\natexlab{a}}.
\newblock URL \url{https://arxiv.org/abs/2508.14153}.

\bibitem[Zhu et~al.(2025{\natexlab{b}})Zhu, Tian, Chen, Zhou, Guo, Liu, Yang,
  and Shen]{zhu2025segagent}
Muzhi Zhu, Yuzhuo Tian, Hao Chen, Chunluan Zhou, Qingpei Guo, Yang Liu, Ming
  Yang, and Chunhua Shen.
\newblock Segagent: Exploring pixel understanding capabilities in mllms by
  imitating human annotator trajectories.
\newblock \emph{arXiv preprint arXiv:2503.08625}, 2025{\natexlab{b}}.
\newblock URL \url{https://arxiv.org/abs/2503.08625}.

\bibitem[Zou et~al.(2022)Zou, Dou, Yang, Gan, Li, Li, Dai, Behl, Wang, Yuan,
  Peng, Wang, Lee, and Gao]{zou2022xdecoder}
Xueyan Zou, Zi-Yi Dou, Jianwei Yang, Zhe Gan, Linjie Li, Chunyuan Li, Xiyang
  Dai, Harkirat Behl, Jianfeng Wang, Lu~Yuan, Nanyun Peng, Lijuan Wang,
  Yong~Jae Lee, and Jianfeng Gao.
\newblock Generalized decoding for pixel, image, and language, 2022.
\newblock URL \url{https://arxiv.org/abs/2212.11270}.

\bibitem[Zou et~al.(2023)Zou, Yang, Zhang, Li, Li, Wang, Wang, Gao, and
  Lee]{zou2023seem}
Xueyan Zou, Jianwei Yang, Hao Zhang, Feng Li, Linjie Li, Jianfeng Wang, Lijuan
  Wang, Jianfeng Gao, and Yong~Jae Lee.
\newblock Segment everything everywhere all at once.
\newblock In \emph{Thirty-seventh Conference on Neural Information Processing
  Systems}, 2023.
\newblock URL \url{https://openreview.net/forum?id=UHBrWeFWlL}.

\end{thebibliography}
